\documentclass[lettersize,journal]{IEEEtran}

\usepackage{amsmath,amsfonts}
\usepackage{algorithmic}
\usepackage{array}
\usepackage[caption=false,font=normalsize,labelfont=sf,textfont=sf]{subfig}
\usepackage{textcomp}
\usepackage{stfloats}
\usepackage{svg}
\usepackage{url}
\usepackage{verbatim}
\usepackage{graphicx}
\usepackage{makecell}
\def\BibTeX{{\rm B\kern-.05em{\sc i\kern-.025em b}\kern-.08em
    T\kern-.1667em\lower.7ex\hbox{E}\kern-.125emX}}
\usepackage{balance}

\begin{document}
\newcommand{\microns}{$\mu$m}

\title{OpenContrails: Benchmarking Contrail Detection on GOES-16 ABI}

\author{Joe Yue-Hei Ng, Kevin McCloskey, Jian Cui, Vincent R. Meijer, Erica Brand, Aaron Sarna, Nita Goyal, Christopher Van Arsdale, Scott Geraedts
\thanks{Manuscript created March, 2023; 
J. Ng, K. Mccloskey, J. Cui, E. Brand, A. Sarna, N. Goyal, C. Van Arsdale, and S. Geraedts are with the Research Department, Google Inc. (e-mail: yhng@google.com). V. R. Meijer is with the Laboratory for Aviation and the Environment, Massachusetts Institute of Technology.
}}

\maketitle

\begin{abstract}
Contrails (condensation trails) are line-shaped ice clouds caused by aircraft and are likely the largest contributor of aviation-induced climate change. Contrail avoidance is potentially an inexpensive way to significantly reduce the climate impact of aviation. An automated contrail detection system is an essential tool to develop and evaluate contrail avoidance systems. In this paper, we present a human-labeled dataset named \emph{OpenContrails} to train and evaluate contrail detection models based on GOES-16 Advanced Baseline Imager (ABI) data. We propose and evaluate a contrail detection model that incorporates temporal context for improved detection accuracy. The human labeled dataset and the contrail detection outputs are publicly available on Google Cloud Storage at \emph{\texttt{gs://goes\_contrails\_dataset}}.
\end{abstract}


\section{Introduction}

Contrails, or condensation trails, are cirrus clouds created by aircraft when flying through cold and humid regions. By trapping infrared radiation which would have otherwise escaped into space, these contrails warm the earth. Recent works have suggested that the warming produced by global aviation through these contrails is comparable to the warming produced through CO$_2$ emissions  \cite{lee2021contribution, bickel2020estimating}. 
Studies have suggested that the majority of contrail warming can be attributed to a small fraction of flights, and their warming impact can be significantly reduced by using a different route \cite{avila2019reducing,teoh2020mitigating}. However considerable uncertainties exist both around the overall impact of contrails \cite{lee2021contribution}, and how well their formation and persistence can be predicted in advance \cite{gierens2020}. More complete knowledge of where contrails are actually forming is key to reducing this uncertainty, and this can be obtained by creating an automated contrail detector with high spatio-temporal coverage.

\begin{figure}[h!]
\begin{center}
    \vspace{2px}
    \includegraphics[width=0.76\linewidth]{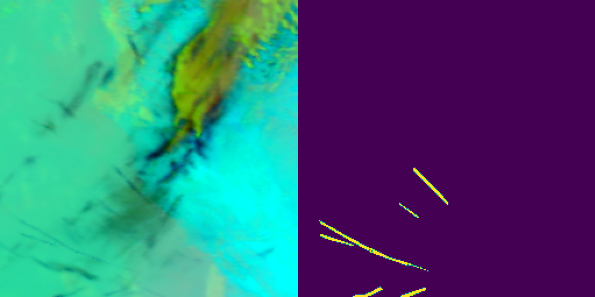} \\
    \vspace{2px} 
    \includegraphics[width=0.76\linewidth]{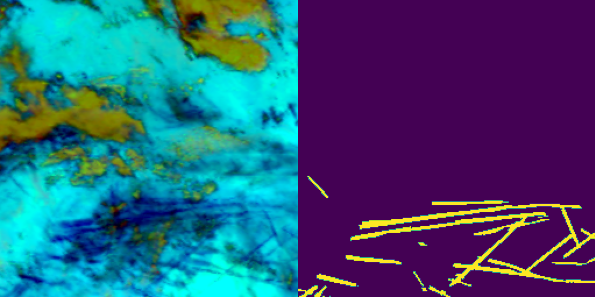} \\
    \vspace{2px}
    \includegraphics[width=0.76\linewidth]{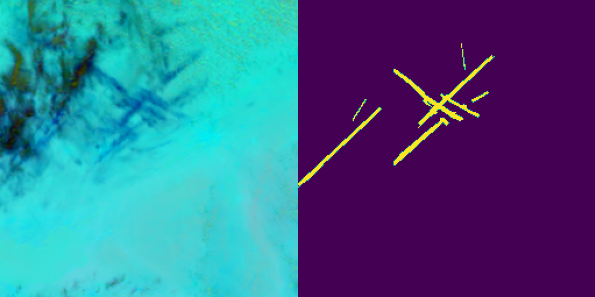} \\
    \vspace{2px}
    \includegraphics[width=0.76\linewidth]{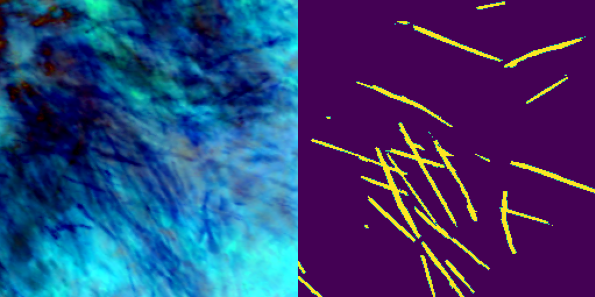}
    \caption{Example images and labels in the dataset. Left: False color images from GOES-16 satellite; Right: Human labeled binary contrail mask.}
    \label{fig:example}
\end{center}    
\end{figure}

Detecting contrails from satellite images is challenging because of their visual similarity to natural cirrus. Contrails are formed as line-shaped ice cloud and then slowly deform over time, becoming difficult to distinguish from natural cirrus. In this work we focus on detecting young contrails that are still linear in shape.
While computer vision has made substantial progress thanks to the rise of large scale datasets, like ImageNet~\cite{deng2009imagenet}, and the use of neural networks, recognition on domain specific problems like infrared satellite images are still challenging due to the lack of large in-domain datasets for pre-training. Apart from that, unlike traditional object recognition tasks where textures and colors are informative, contrails often look similar to or even indistinguishable from natural cirrus. In addition to color and texture, human experts must take into account the overall shape of the contrails (linear) and their temporal evolution (unlike natural cirrus, contrails appear very quickly and spread out rapidly over time).

Challenging computer vision tasks have often benefited from the existence of high-quality datasets, such as the MNIST dataset~\cite{mnist} for handwriting recognition. Such datasets lower the barrier to entry for different research groups to try to solve the problem, and provide a standard way to compare new models to prior work to assess whether a given new technique has led to improvement. In this work we provide a public dataset, \emph{OpenContrails}, to facilitate model comparison and reproducible research.

While previous works have attempted to train contrail detection models with deep convolutional neural networks~\cite{meijer2022contrail,zhang2018contrail}, the performance of those models are still significantly behind that of a human expert, and therefore improvement in model performance is possible. One possible reason why humans can do better is that during labelling, humans can evaluate the temporal behavior of the contrails, whereas existing computer models perform detection based on only one frame. We develop a model that takes multiple input frames into account for temporal context. We show that such a model improves the the detection performance when compared to models that only process a single frame.

In this work, we present a dataset of human-labeled contrail detections on the geostationary satellite GOES-16, which provides high spatio-temporal coverage, and we make it publicly available. The proposed \emph{OpenContrails} dataset contains high quality per-pixel human label contrail masks for each image in the dataset. We train a neural network for detecting contrails and show that the resulting model produces high quality contrail detection outputs. We run the contrail detection model on multiple years of available GOES-16 images and confirm previous findings on contrail research, such as contrail coverage patterns and diurnal effects. We envision the \emph{OpenContrails} dataset as the benchmark for the contrail detection task, and hope it accelerates contrail research by making large-scale contrail detection results widely available.

\subsection{Related Work}
The effect of contrails on climate change has been widely studied in previous literature \cite{chen2013simulated,schumann2012contrail,bickel2020estimating,lee2021contribution}.  Simulations indicate that a small set of flights are responsible for the majority of contrail warming, which suggests that it is possible to adjust a small number of flight routes to substantially reduce global warming~\cite{avila2019reducing,teoh2020mitigating}.

Empirical analyses of contrails from satellite imagery in the last two decades~\cite{mannstein1999operational,meyer2002regional,vazquez2015contrail} were typically accomplished by the algorithm proposed by Mannstein et al \cite{mannstein1999operational}. Mannstein et al. proposed to detect contrails by applying a set of filters and thresholds on brightness temperature imagery.
Zhang et al. trained convolutional neural networks to detect contrails on Himawari-8 satellite images~\cite{zhang2018contrail} for evaluating contrail coverage. Meijer et al. also trained a neural network for contrail detection on GOES-16 images \cite{meijer2022contrail} to examine what effects the reduced flight traffic during the COVID-19 pandemic had on contrails. None of these works made the dataset or contrail detections publicly available, and while Meijer et al. only labeled the continental US (CONUS) region, we sample scenes to be labeled from the majority of the GOES-16 full-disk viewing extent.

McCloskey et al. released a contrail dataset on Landsat-8 data \cite{mccloskey2021human}. With 30 meter pixels in its infrared channels, Landsat-8 images have sufficient spatial resolution to identify even very young contrails. However due to its low-earth sun-synchronous orbit and mission goals, imagery is only available for scenes that are mostly land, mostly in the daytime, at fixed local times of day and with a 16-day repeat cycle. Thus Landsat-8 imagery has clear limits for conducting large scale analysis for contrail research. We construct our dataset on GOES-16 ABI images~\cite{schmit2017closer} taken from geostationary orbit, which nominally have 2km pixels in the infrared channels but have much higher spatio-temporal coverage. Consequently, our dataset and contrail detection model can be used as a foundation for contrail warming impact assessment and validating contrail avoidance experiments in the western hemisphere.

Deep neural networks have been extensively used for remote sensing applications~\cite{ma2019deep,zhu2017deep}. 
The task of contrail detection can be cast as image segmentation, on which deep neural network have shown great success. Here we employ promising image segmentation techniques including using the ResNet backbone~\cite{he2016deep} and DeeplabV3+~\cite{chen2017rethinking,chen2018encoder} architectures.

\section{Dataset}

We build our dataset using GOES-16 Advanced Baseline Imager (ABI) imagery~\cite{schmit2017closer}, specifically brightness temperatures calculated from the Level 1B radiances using the Planck constants provided by~\cite{definition2019users}. GOES-16 views the North and South American region with a full-disk image taken every 10 minutes since April 2019 (15 minute interval before April 2019). The nominal pixel size of GOES-16 ABI is $2 \times 2$km at nadir. This relatively coarse resolution means that contrails cannot be seen when they initially form, instead we must wait some time for them to spread out enough to be visible. This is not necessarily a disadvantage: the warming effects of contrails are dominated by contrails which persist for hours~\cite{teoh2020mitigating}, so the inability to detect shorter-lived contrails may not hinder our ability to assess and prevent the vast majority of contrail warming.

We generate the examples by randomly sampling image patches from the GOES-16 viewable extent from April 2019 to April 2020. To avoid large viewing angles, we restrict the sampling of image patches to be between -50 to 50 degree latitude and -135 to -30 degree longitude. 
Uniform sampling of GOES-16 images would result in very few positive examples, as contrails occur infrequently. To increase the number of positive examples in the dataset, we downsampled regions unlikely to contain contrails. We did this in three ways. First, we obtained aircraft flight tracks obtained from terrestrial ADS-B data licensed for publication from FlightAware, LLC (\url{https://flightaware.com}). We advected these flight tracks according to European Centre for Medium-Range Weather Forecasts (ECMWF) ERA5~\cite{hersbach2020era5} wind data, using the Runge-Kutta method \cite{bogacki19893}, for four hours. We kept only $5\%$ of the images which contained no advected flight tracks, and only $20\%$ of the images which contained less than $10$ tracks. Second, we calculated the relative humidity over ice from ECMWF ERA5 temperature, pressure, and specific humidity values. Since contrails can only persist if this humidity is above $100\%~$\cite{gierensbook}, we kept only $5\%$ of images for which no location in the image had relative humidity $>90\%$ for any of the altitudes where contrails might form ($7000-12000m$). Finally, we applied the Mannstein et al.~\cite{mannstein1999operational} contrail detection algorithm tuned as in Mccloskey et al. \cite{mccloskey2021human} to achieve very high recall, as opposed to high precision (\cite{mccloskey2021human} showed that a similar method applied to images of contrails in Landsat-8 detected every contrail, but for every contrail it detected it also detected around $20$ false positives.) We kept only $5\%$ of the images that failed this screen. Note that the FlightAware data does not contain all flights and the ECMWF wind and humidity data are not always accurate. Therefore, the filters are not expected to be perfectly accurate, and it is still necessary to have human labelers annotate the images.

To further boost the number of positives in the dataset, we also included some GOES-16 ABI imagery at locations in the US where Google Street View images of the sky contained contrails. To define when a Street View image of the sky contained a contrail, we used 64-dimensional image feature vectors derived from image-text data, created with an approach similar to that used by Juan et al.~\cite{juan2019graph}. We applied a threshold to the cosine similarity of the Street View image feature vector and that of a seed image of a contrail taken from the ground; if it was similar enough, GOES-16 imagery at that time and location were sampled for human labeling of contrails.  These additional labeled images are only in the training set: because Street View cars operate on days with sunnier weather, it may be easier than usual to identify contrails in the GOES-16 imagery of those locations. We found in our experiment that these additional training data slightly improve the contrail detection model performance on the validation set. Note that there are no Street View images in the dataset or used in models reported here: they were only used to identify contrail-rich GOES-16 scenes for human labeling.

\begin{figure}
    \centering
    \includegraphics[width=.4\linewidth]{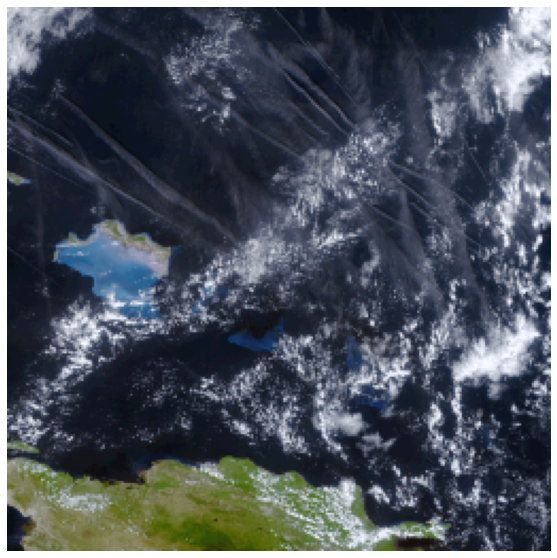}
    \includegraphics[width=.4\linewidth]{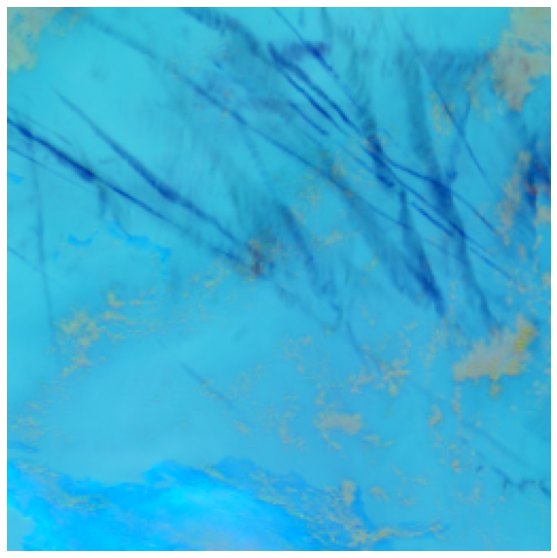}
    \caption{Left: True color image; Right: false color image in ash color scheme.}
    \label{fig:color_example}
\end{figure}

We want to detect contrails throughout both day and night, so we show imagery to human labelers in an ``ash'' false color scheme that combines three longwave GOES-16 brightness temperatures \cite{kulik2019satellite}. The red, blue and green channels are represented by the 12\microns{}, difference between 12\microns{} and 11\microns{}, and difference between 11\microns{} and 8\microns{} respectively. This color scheme is chosen to help identify contrails by highlighting ice-clouds as darker colors. An example of the ash color scheme and the true color RGB image is shown in Figure~\ref{fig:color_example}. Each image patch in the dataset corresponds to an approximately $500 \times 500$ km region and is reprojected to the Universal Transverse Mercator (UTM) coordinate system (with the zone decided by the northwest corner of the region) using bilinear resampling. Labelers are shown image patches with size of $281\times 281$. As the labels near the boundary are often noisy due to the lack of spatial context, we center crop to $256 \times 256$ pixels on the images and contrail masks for training and evaluation. 

\begin{figure}
    \centering
    \includegraphics[width=.78\linewidth]{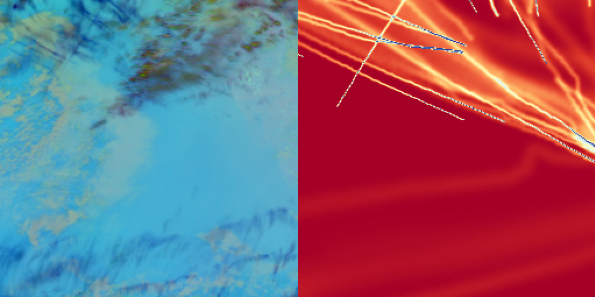}
    \caption{Example GOES-16 false color image with advected flight density shown to labelers.}
    \label{fig:goes_with_flight_density}
\end{figure}

\begin{figure*}
    \centering
    \includegraphics[width=.125\textwidth]{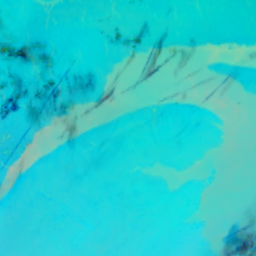}\includegraphics[width=.125\textwidth]{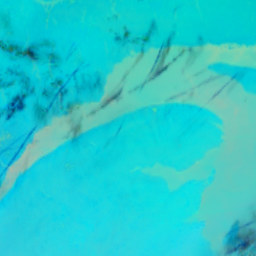}\includegraphics[width=.125\textwidth]{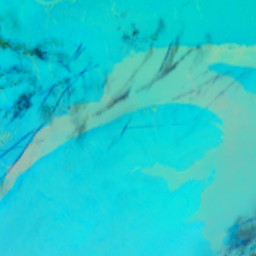}\includegraphics[width=.125\textwidth]{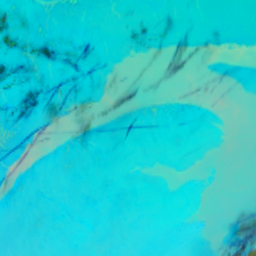}\includegraphics[width=.125\textwidth]{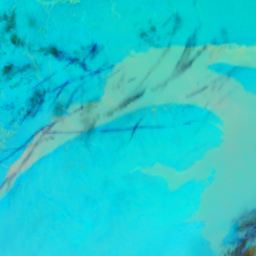}\includegraphics[width=.125\textwidth]{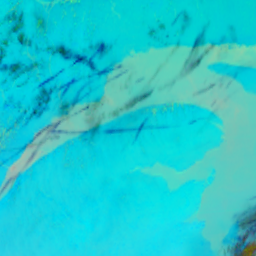}\includegraphics[width=.125\textwidth]{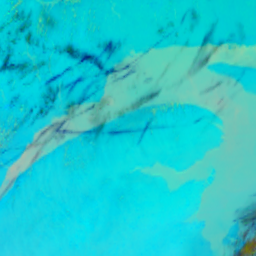}\includegraphics[width=.125\textwidth]{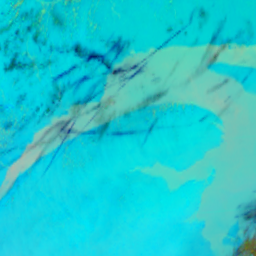}
    \caption{Example sequence showing evolution of contrails. Contrails appear quickly and start as straight lines. They deform and fade over time slowly from satellite observations.}
    \label{fig:contrail_evolution}
\end{figure*}

Since the temporal context is important for recognizing contrails, we show labelers 5 images (50 minutes) before and 2 images (20 minutes) after the image being labeled. An example image sequence of contrail evolution is shown in Figure~\ref{fig:contrail_evolution}, which shows how contrails appear spontaneously and spread out over time, while natural cirrus often occur for a longer time and may get sharper over time instead of spreading out. The advected flight density is also shown to the labelers to assist labeling. It is generated from FlightAware flight data and is advected using the same method as for screening (ECMWF ERA5 wind vector data, with the Runge-Kutta method). We spread out the advected flight density of flights which are older (at the time the image was taken) to reflect how uncertainty in the wind vector data makes us less certain about where a contrail will ultimately advect to. We ask the labelers to still label the contrails if flights are not shown in the flight density only when they are very confident. An example GOES-16 ``ash'' image and the corresponding advected flight density is shown in Figure~\ref{fig:goes_with_flight_density}.

We ask human labelers to draw polygons on the images to enclose the contrail pixels in a web-based labeling interface. 
The labelers are instructed on how to label contrails through a set of slides containing guidelines, including examples of contrails and non-contrail cirrus.

We provide detailed instructions for labelers to identify and label contrails, for example: contrails are typically seen as dark linear objects in the satellite images using the ash color scheme; contrails appear as sharp lines and then spread out over time; and static structures across multiple frames are likely to be ground based objects such as roads, rivers or coastlines.
We also provide the following additional labeling guidelines:
\begin{enumerate}
    \item Contrail must contain at least 10 pixels to be labeled; 
    \item At some time in their life, contrails must be at least 3x longer than their width;
    \item Contrails must either appear suddenly or enter from the sides of the image;
    \item Contrails should be visible in at least two images.
\end{enumerate}

Labelers were asked to initially label a small set of images, and then authors of this work gave feedback on their initial label quality. The process iterates until a labeler generates labels with satisfactory accuracy. As we observe variance across individual labelers, each GOES-16 scene is labeled by 4 human labelers, and pixels are considered positive if at least 3 labelers considered the pixel a contrail.

Even when instructions and trainings are provided, identifying and labeling contrails on satellite images is still challenging. We compared the labeler performance compared to the groundtruth (labeled by authors of this work) on about 200 examples. Individual labelers give about 64\% precision and 76\% recall measured per-pixel. The aggregated labels from 4 labelers with majority vote gives about 80\% precision and 78\% recall.

\begin{figure}
    \centering
    \includegraphics[width=.99\linewidth]{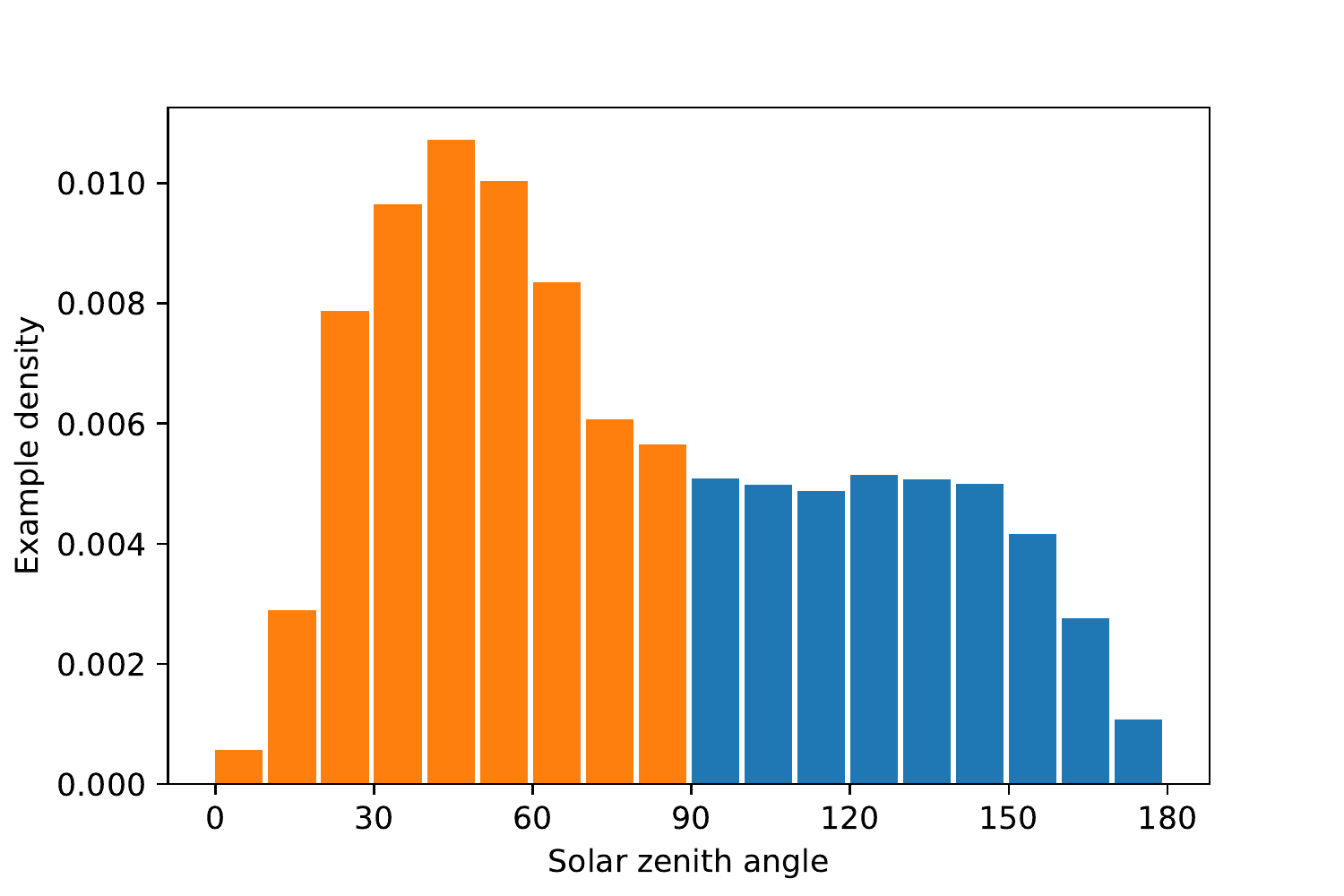}
    \caption{Distribution of solar zenith angle in the dataset. Orange and blue bars correspond to daytime and nighttime examples respectively.}
    \label{fig:zenith_distribution}
\end{figure}

\begin{figure}
    \centering
    \includegraphics[width=.99\linewidth]{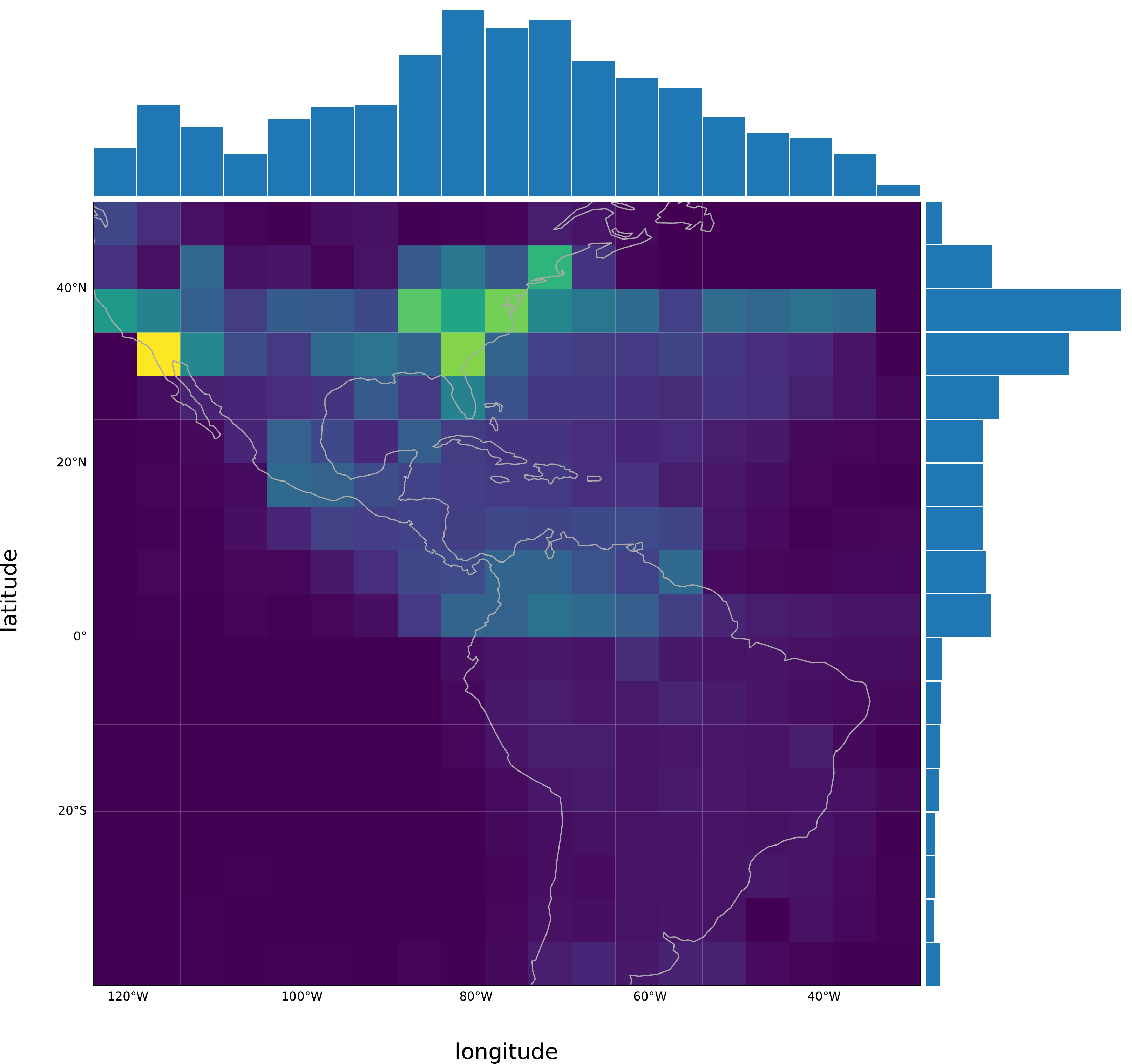}\\
    \vspace{5px}
    \includegraphics[width=.7\linewidth]{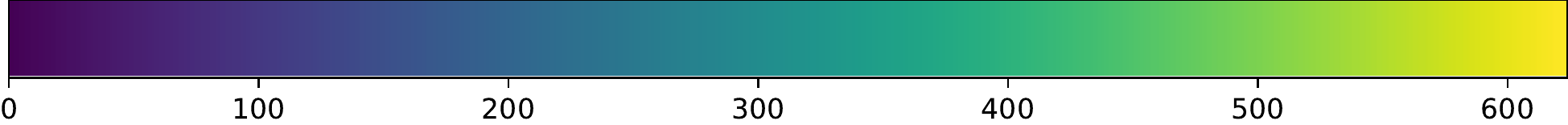}
    \caption{Spatial distribution according to the image center in the dataset. The top and the right histograms correspond to the marginal distributions along the latitude and longitude axes.}
    \label{fig:distribution}
\end{figure}

The full dataset contains 20,544 examples in the train set and 1,866 examples in the validation set. The examples are randomly partitioned except for the satellites scenes that were identified as likely to have contrails by Google Street View, which are only included in the training set. 9,283 of the training examples contain at least one annotated contrail. About 1.2\% of the pixels in the training set are labeled as contrails. 
The dataset contains a wide variety of times and locations, as shown in Figure \ref{fig:zenith_distribution} and \ref{fig:distribution}. The examples are not uniformly distributed in space and time as the images are sampled to include more contrail examples as described above. 

\begin{figure}
    \centering
    \includegraphics[width=.83\linewidth]{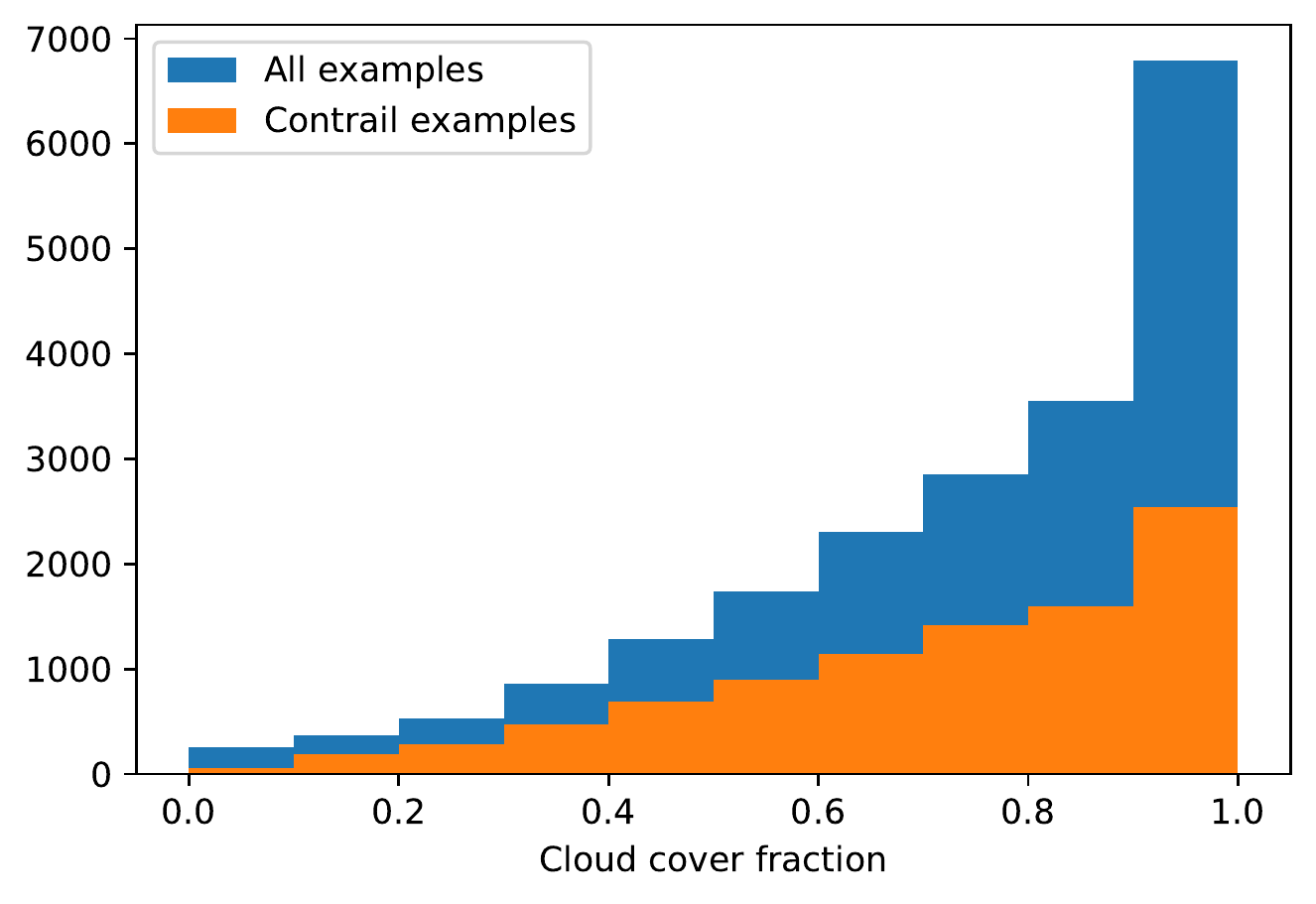}
    \caption{Statistics of cloud coverage in the dataset.}
    \label{fig:dataset_cloud_cover}
\end{figure}

\begin{figure}
\centering
\vspace{-1em}
\includegraphics[width=.96\linewidth]{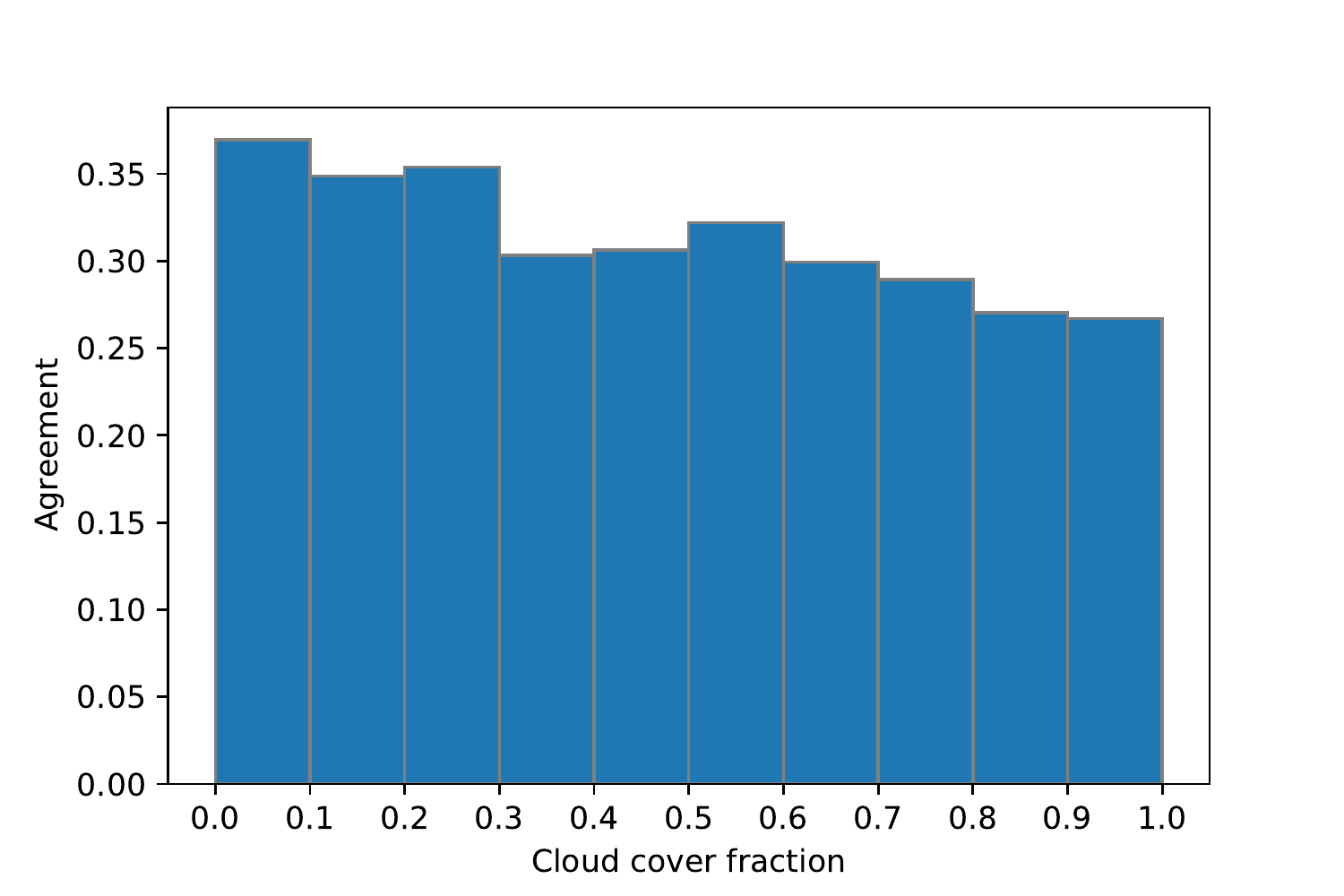}
\caption{Labeler agreement at different cloud cover fractions.}
\label{fig:dataset_cloud_cover_agreement}
\end{figure}

Contrails occur more often in cloudy scenes. We estimate the cloud cover fraction using the GOES-16 ABI L2 Cloud Top Phase product and regard all non-clear-sky pixels as clouds. We show the example counts in our dataset by the cloud coverage fraction in Figure \ref{fig:dataset_cloud_cover}. Our dataset contains more cloudy scenes than clear sky images, and the positives examples are roughly proportionally distributed at different cloud coverage fractions. It is possible that contrails are harder to be identified in cloudy scenes. Figure~\ref{fig:dataset_cloud_cover_agreement} shows the labeler agreement at different cloud cover fractions in the scenes, where agreement is defined as the number of contrail pixels labeled by the majority of the labelers divided by the number of contrail pixels labeled by at least one labelers. Labelers agree slightly less in scenes with higher cloud cover fractions.

The dataset is provided in TFRecord format. Each example includes an image with size of $256\times 256$. Each example is provided with multiple GOES-16 ABI brightness temperatures and brightness temperature differences. The dataset will be uploaded to Google Cloud bucket and made publicly available.

\subsection{Evaluation Metrics}
The best evaluation metric for contrail detection depends on the downstream application. A per-pixel metric would be useful for assessing the energy forcing of contrails as it captures the area covered by contrails, and a per-contrail metric would be more useful for associating the contrails to flights and computing the contrail lengths. Here we suggest two evaluation metrics for contrail detection.

Contrail detection can be considered as an image segmentation task which is traditionally evaluated with per-pixel metrics. Typical semantic segmentation metrics like intersection-over-union (IoU) and DICE coefficient are sensitive to the choice of a single binarization threshold. Since the detection outputs could be useful at different operating points for different applications, we evaluate the per-pixel performance using area under the precision recall curve (AUC-PR). We use 10,000 thresholds sampled uniformly between 0 and 1 to compute the precision and recall for each pixel to obtain the AUC-PR.

We also evaluate contrail detections as a line segment detection task. The contrail detection system should output a set of line segments for each image. Similar to~\cite{mccloskey2021human}, we evaluate the resulting linear contrails with a precision recall curve. The contrail detected is considered a correct detection if its angle is within 10 degrees from the groundtruth and within 10km (approximately 5 pixels) mean distance to the contrail~\cite{mccloskey2021human}.

\section{Models}
We use a convolutional neural network model for contrail detection. The model outputs a score (between 0 to 1) for each pixel indicating the confidence that pixel is part of a contrail.  The per-pixel contrail detection outputs are then postprocessed to produce line segments representing the contrail instances.

To consistently identify contrails including at night we train on infrared channels represented as brightness temperatures as input.
Our contrail detection model takes 6 channels as input: 8\microns{}, 10\microns{}, 11\microns{}, 12\microns{}, difference between 12\microns{} and 11\microns{}, and difference between 11\microns{} and 8\microns{}. Each input channel is standardized by subtracting the global mean and dividing by the global variance of the channel before feeding it into the network.

\subsection{Single-frame Model}
We first train a binary image segmentation model for classifying each pixel as contrail or background. We employ the semantic segmentation architecture DeeplabV3+ \cite{chen2018encoder} and experimented with different image backbones. Following \cite{chen2018encoder}, we use a dilated ResNet~\cite{chen2017rethinking} by modifying the last residual block with dilated convolutions, resulting in an output stride of 16 in the output feature maps. An Atrous Spatial Pyramid Pooling (ASPP) module~\cite{chen2018encoder} follows, to incorporate information at multiple scales. Low level feature maps are then combined with upsampled higher level feature maps to preserve the resolution of the feature maps. The final segmentation head consists of three convolutional layers and a 2-channel output representing `background' or `contrail'. The model is trained with a per-pixel cross-entropy loss.

It has been shown that the performance of the original ResNet can be improved by various recent techniques~\cite{bello2021revisiting}. We follow the best practice documented in \cite{bello2021revisiting} and use an improved architecture with ResNet-D stem~\cite{he2019bag} and Squeeze-and-Excitation blocks~\cite{hu2018squeeze}. We set the squeeze-and-excitation ratio to 0.25. Stochastic depth~\cite{huang2016deep} was used to regularize the network with 0.2 drop rate. 

The output resolution of the segmentation model can be smaller than the input images by a factor of the output stride of the backbone. We apply bilinear upsampling to the network prediction to resize the outputs to the groundtruth image resolution during both training and evaluation.

\subsection{Multi-frame Model}
To incorporate temporal context into our model we extend the DeeplabV3+ model to incorporate multiple input frames. We use a spatio-temporal encoder (based on 3D convolutions) to incorporate temporal information.

Inspired by the wide success of convolutional networks in video classification, we use the inflated 3D (I3D) convnet as the backbone of our detection model. Our inflated convnet is based on the ResNet architecture~\cite{he2016deep}. Since 3D convolution is computationally expensive, we implement the inflated ResNet block by factoring spatial and temporal convolutions similar to ~\cite{wang2018non, qiu2017learning}. A regular ResNet block consists of three convolution layers with $1\times1$, $k\times k$, and $1\times1$ kernel sizes. We inflate each ResNet block by expanding the first convolutional layer with a temporal component, while keeping the second and third convolutional layers with a temporal kernel size of 1. The inflated ResNet block thus has kernel sizes of $t \times 1 \times 1$, $1 \times k \times k$, and  $1 \times 1 \times 1$. In our implementation, we apply the temporal convolution with a kernel size of 3, at every ResNet block in stages 3 and 4.

Differently from video classification (where only a single label is returned for each example), fine-grained spatio-temporal information is required for precisely predicting the contrail detection masks. We augment the backbone network to have a temporal stride of 1 in the first max pooling and convolution layers in the stem. Thus, the temporal resolution is fixed throughout the backbone. 

\begin{figure*}
    \centering
    \includegraphics[width=0.9\textwidth]{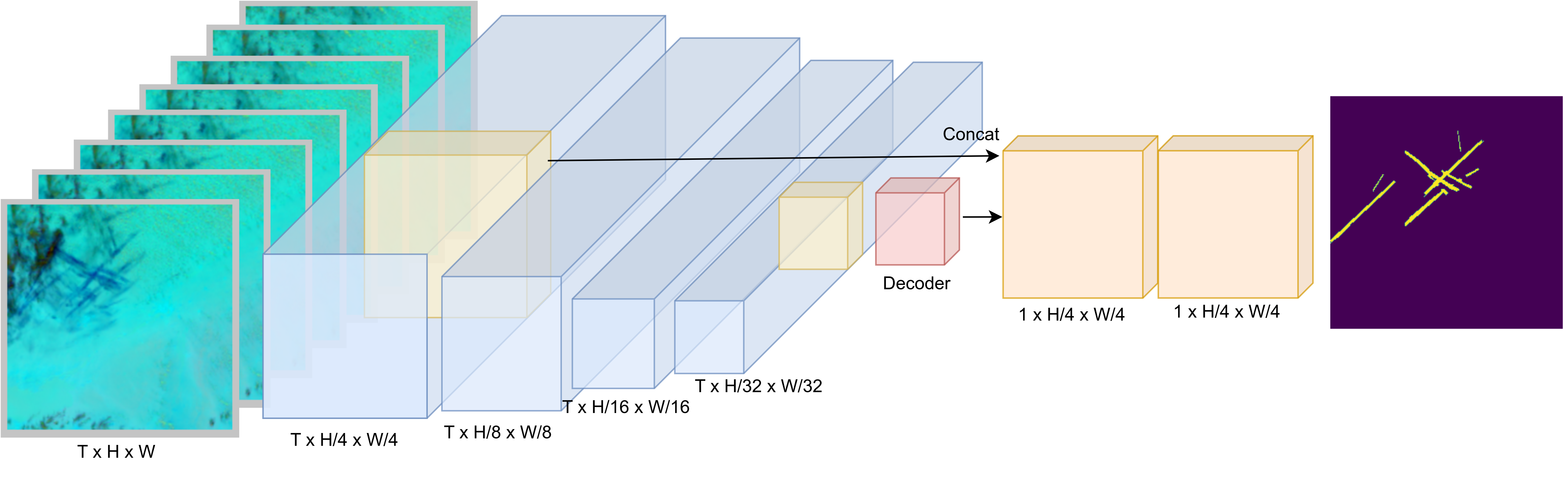}
    \caption{Network architecture for multi-frame detection model. The model takes a sequence of frames as inputs, and take a slice of the feature maps that correspond to a single timestep, and then run the decoder.}
    \label{fig:architecture}
\end{figure*}

Since the dataset only has one labeled frame available for each example, we slice one frame of features to pass to the decoder. Similar to the single-frame model, we use the ASPP module for the decoder and stacked convolutional layers for classifier. The overall model architecture is illustrated in Figure~\ref{fig:architecture}. 

\subsection{Training}
We train the network using stochastic gradient descent with 10,000 iterations, using a batch size of 256 and a momentum of 0.9. All models are implemented in TensorFlow and trained on 16 Tensor Processing Unit V3 (TPUv3) chips. The learning rate is set to 0.2 and gradually decays to zero with half-cycle cosine learning rate decay. To improve training stability, we employ a linear warm up learning rate schedule, increasing the learning rate linearly from 0 for the first 500 iterations. L2 weight decay is used with a factor of 0.0001. Data augmentation is applied during training to reduce overfitting: a scale factor is chosen uniformly randomly at each iteration, and then the image is scaled according to the scale factor, and a crop of the scaled image is used for training. We found this significantly helps overall performance.

As our detection dataset is relatively small compared to natural image datasets like ImageNet~\cite{deng2009imagenet}, pre-training the neural network on larger datasets can often help --- even pre-training on images which are very different from satellite images. We initialize the network with ImageNet pre-trained weights for the single-frame model. We have also experimented with weights pre-trained on image segmentation datasets like COCO but did not see improvement. For the multi-frame model, we pre-train the backbone model on the Kinetics-400 dataset~\cite{kay2017kinetics} which consists of about 240,000 videos collected from YouTube: each video clip is 10 seconds long and is annotated with an action from 400 classes.

\subsection{Converting binary masks to line segments}
\label{sec:linearize}
The network outputs corresponding to pixel-wise contrail probabilities are further post-processed to produce the line segment outputs. We first choose a threshold to convert the pixel probabilities to binary masks, and then use off-the-shelf OpenCV's LineSegmentDetector, which is an implementation of~\cite{von2012lsd}, to detect line segments from the binary masks. We then iteratively merge line segments if one segment is part of (or a continuation of) another line segment. We found that this procedure works similarly well compared to a more complicated algorithm which convolves multiple line kernels at multiple scales as described in  \cite{mannstein1999operational,mccloskey2021human}, which requires tuning about 10 hyperparameters for the best performance.

\section{Results}

\subsection{Main Results}

We compare the single frame models and multi-frame model with different backbones. The results are summarized in  Table~\ref{tab:backbone}.
All models achieve reasonable detection performance, showing that our dataset is sufficiently large for training contrail detection models. The multi-frame model slightly outperforms the single-frame based models, showing the model is able to use temporal context to improve detection.

\begin{table}[ht]
\caption{Different architectures for contrail detection.}
\label{tab:backbone}
\centering
\begin{tabular}{c|c}
     Backbone & Pixel AUC-PR \\
     \hline
     ResNet-50 & 68.9 \\
     ResNet-101 & 69.5 \\
     Dilated ResNet-101 & 70.5 \\
    \hline
    ResNet3D-101 & \textbf{72.7} \\
\end{tabular}
\label{tab:results_architecture}
\end{table}

\begin{figure}
    \centering
    \includegraphics[width=0.95\linewidth]{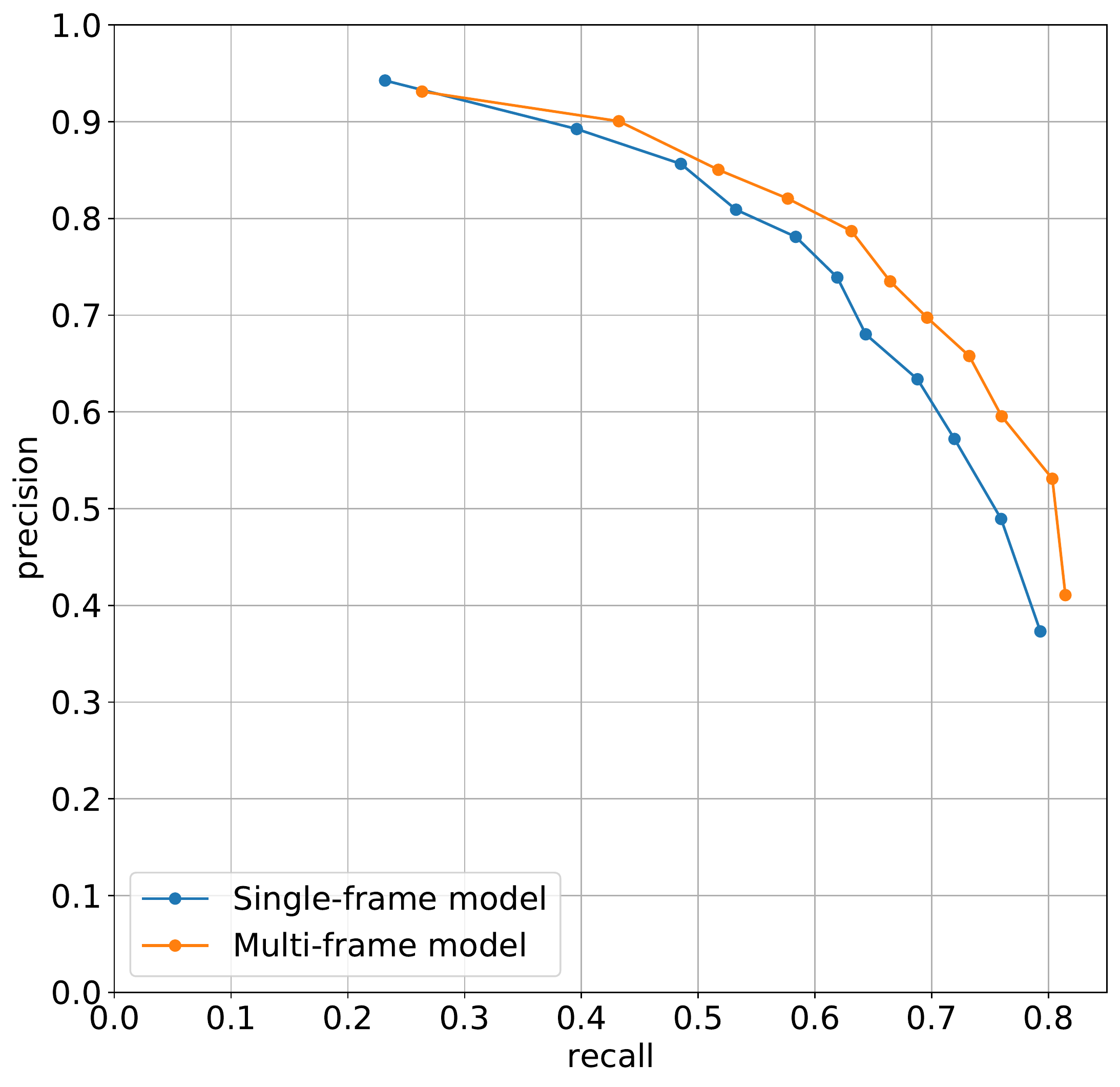}
    \caption{Per-contrail precision/recall curves, showing better precision and recall for the multi-frame model compared to the best single-frame model we explored.}
    \label{fig:pr_curve}
\end{figure}

\textbf{Linear Contrail Detection.}
To assess model accuracy on linear contrail objects, we linearize the output prediction mask as discussed in Section \ref{sec:linearize}. Before each linearization, we vary the thresholding value for the model output mask to generate multiple binary contrail masks at different confidence levels. Figure~\ref{fig:pr_curve} shows the precision recall curves from this process. Once again our multi-frame model shows better performance compared to the single-frame model.

McCloskey et al. \cite{mccloskey2021human} provided a similar precision recall curve on the contrail dataset based on the Landsat-8 satellite with the baseline Mannstein et al algorithm~\cite{mannstein1999operational}. It is worth noting that our precision presented here is significantly higher. At 60\% recall, The Mannstein et al. algorithm gave 15\% precision on the Landsat-8 dataset, while our model gives more than 70\% precision on our GOES-16 based dataset. While the evaluation datasets and models are different, this suggests that our model detects contrails more accurately for contrail research.

\begin{figure}
    \centering
    \includegraphics[width=0.9\linewidth]{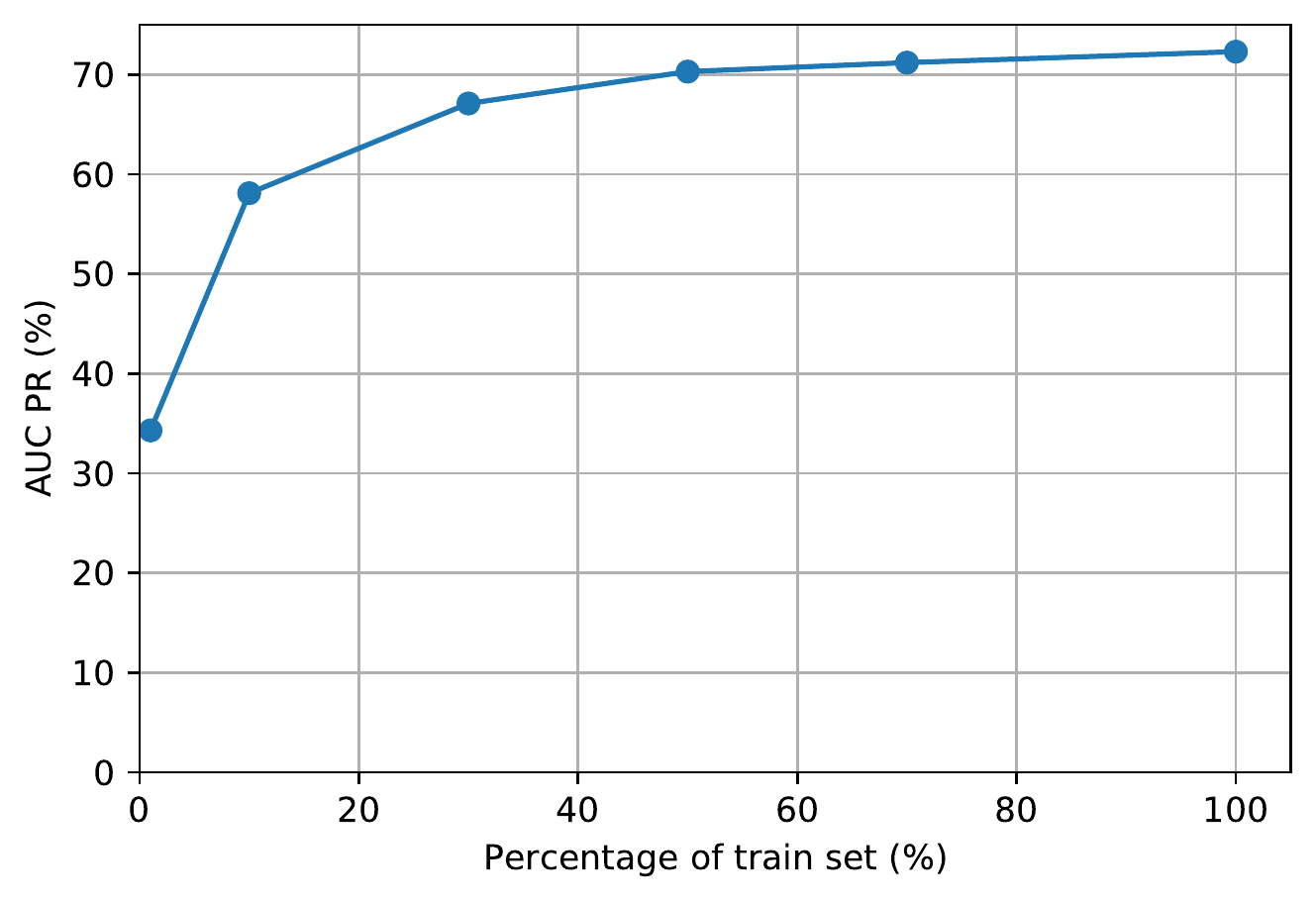}
    \caption{Validation performance (pixel-wise AUC-PR) when the multi-frame model is trained with different numbers of labels.}
    \label{fig:train_set_size}
\end{figure}

\subsection{Detailed analyses}

\textbf{Input image size.} Using a larger input image size often improves performance on visual recognition tasks~\cite{tan2019efficientnet,lin2017focal}. This is because the network can use more computation on smaller patches, and the output resolution is increased. We perform bilinear resampling to enlarge the input image before feeding it to the network, during both training and evaluation. The output predictions are resampled to match the groundtruth resolution when computing the AUC-PR. Table~\ref{tab:resolution} shows the results with different resolution. We see that larger input image sizes give slightly better results, at the cost of slower inference and more memory usage during training.
\begin{table}[ht]
\centering
\caption{Contrail detection (multi-frame model) with different input resolutions.}
\label{tab:resolution}
\begin{tabular}{c|c|c}
     Input resolution & \makecell{Single-frame Model \\ Pixel AUC-PR} & \makecell{Multi-frame Model \\ Pixel AUC-PR} \\
     \hline
     256x256 & 63.2 & 65.1\\
     384x384 & 68,7 & 70.7\\
     512x512 & 70.5 & \textbf{72.7} \\
     784x784 & 71.5 & -\\
     1024x1024 & 71.8 & - \\
\end{tabular}
\end{table}

\textbf{Number of input frames.} Using multiple input frames improves the performance by providing more temporal context. The results are summarized in Table~\ref{tab:input_frames}. In general, we observe that the temporal context before the labeled frames is more important than after the labeled frames. This matches our intuition as contrails dissipate over time and become harder to recognize.
\begin{table}[ht]
\centering
\caption{Detection performance with different numbers of input frames.}
\begin{tabular}{c|c}
     Size of temporal context & Pixel AUC-PR \\
     \hline
     $T_{\mathrm{before}}=0, T_{\mathrm{after}}=0$ & 71.4 \\
     $T_{\mathrm{before}}=0, T_{\mathrm{after}}=1$ & 71.3 \\
     $T_{\mathrm{before}}=1, T_{\mathrm{after}}=0$ & 71.7 \\
     $T_{\mathrm{before}}=0, T_{\mathrm{after}}=3$ & 70.4 \\
     $T_{\mathrm{before}}=2, T_{\mathrm{after}}=1$ & 71.7 \\
     $T_{\mathrm{before}}=3, T_{\mathrm{after}}=0$ & \textbf{72.7} \\
     $T_{\mathrm{before}}=4, T_{\mathrm{after}}=3$ & 72.0 \\
     
\end{tabular}
\label{tab:input_frames}
\end{table}

\begin{figure}
    \centering
    \includegraphics[width=\linewidth]{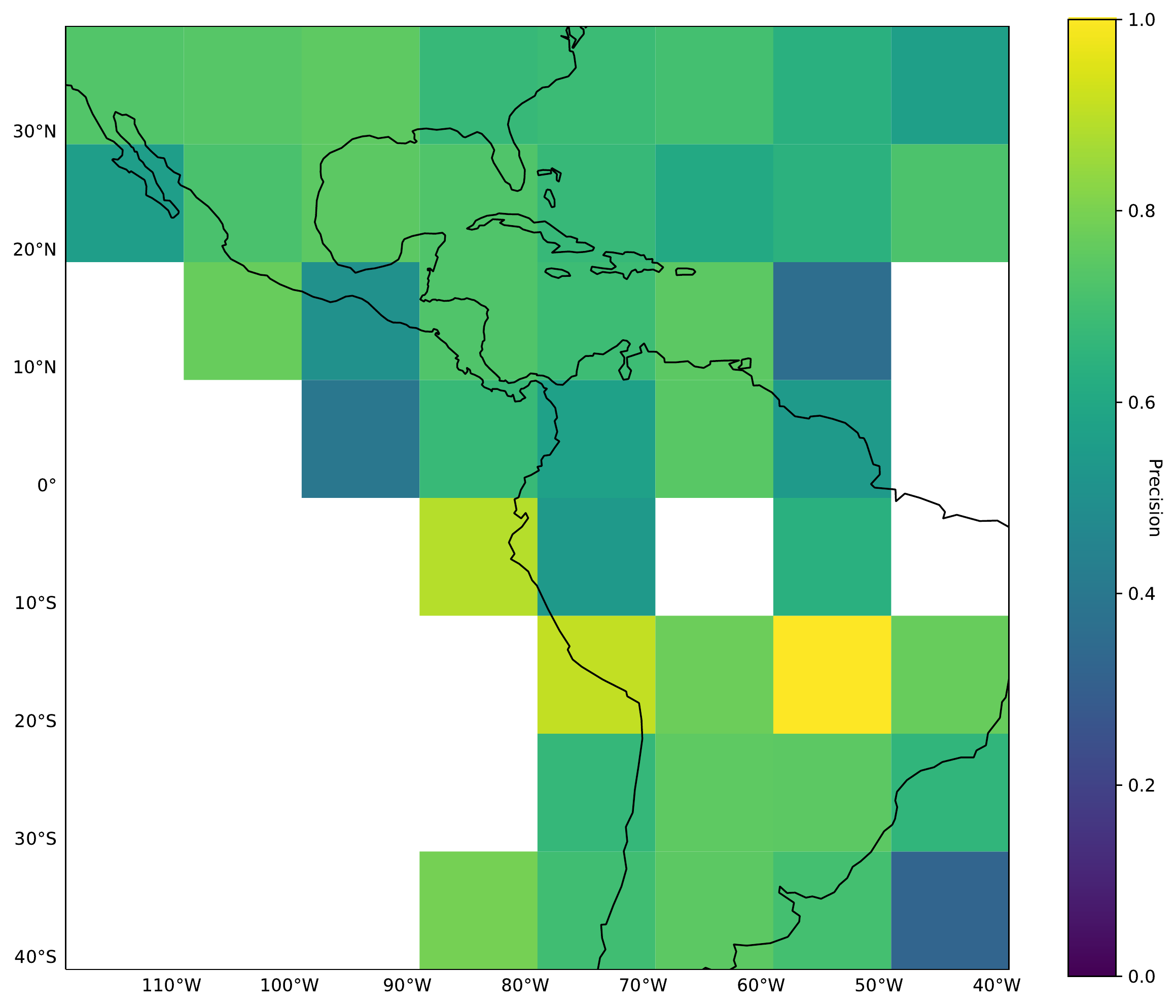}
    \includegraphics[width=\linewidth]{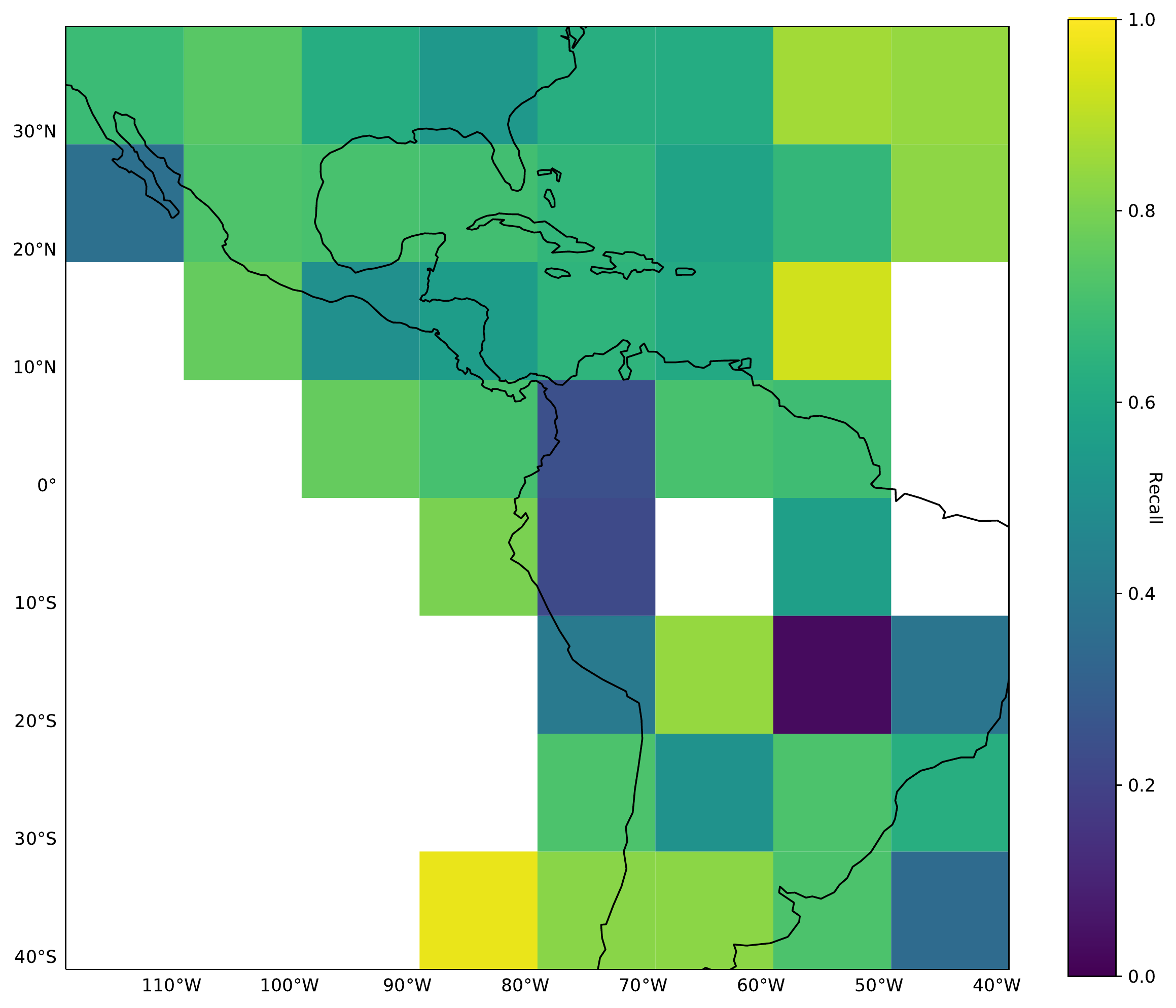}
    \includegraphics[width=\linewidth]{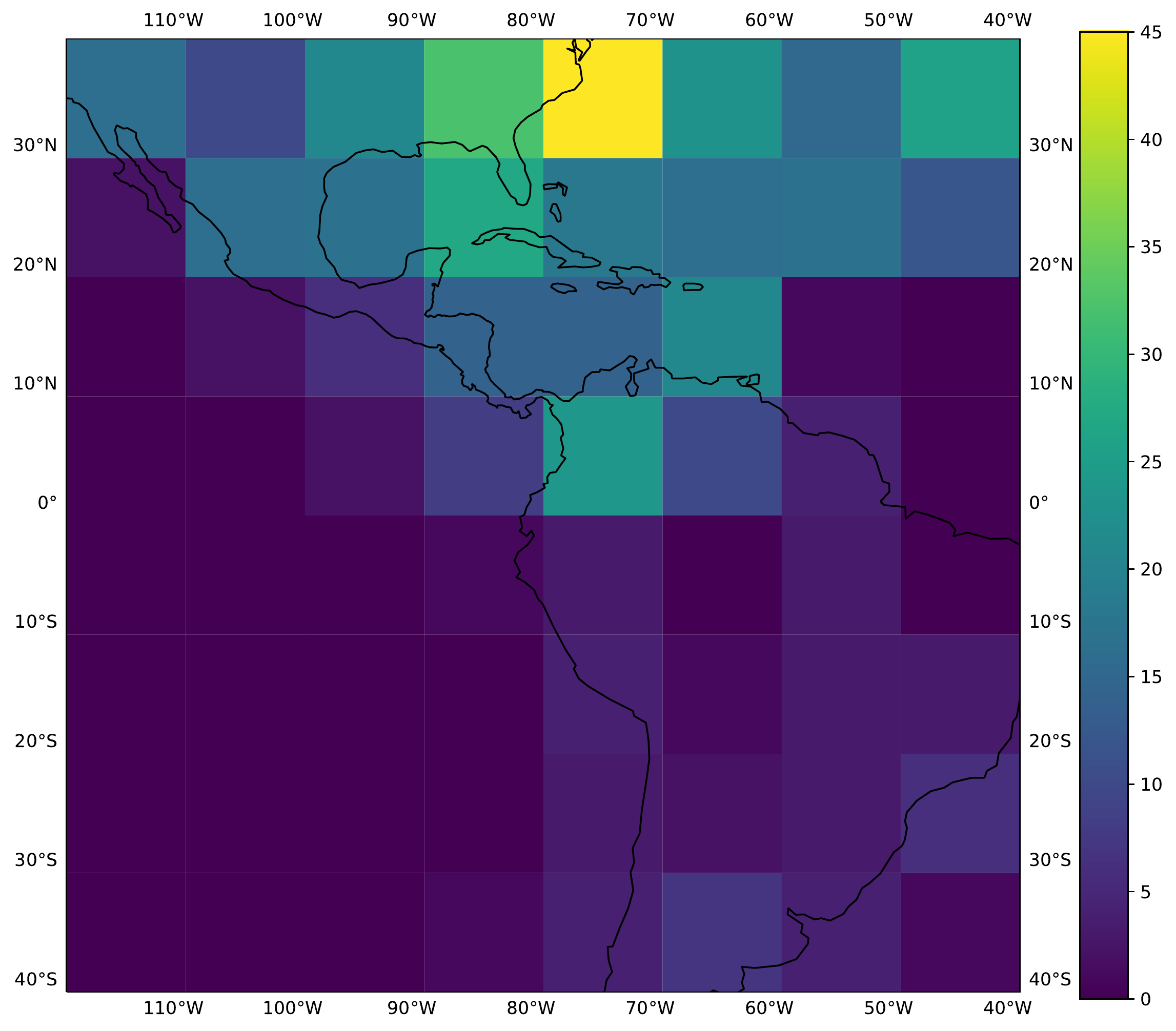}
    \caption{Detector precision (top), recall (middle), and number of positive examples (bottom) at different spatial locations.}
    \label{fig:spatial_performance}
 \end{figure}

\textbf{Performance across spatial locations.} We visualize the precision and recall (with 0.4 detection threshold) at different spatial locations in Figure~\ref{fig:spatial_performance} for each 10x10 degree gridbox that contains at least one positive examples in our validation set. The precision and recall estimates are noisier in South America due to smaller number of positive examples. In general, we observe no substantial spatial bias in different spatial locations, except with slightly worse performance in the northwest part of South America.

\begin{figure*}
    \centering
    \includegraphics[width=.32\textwidth]{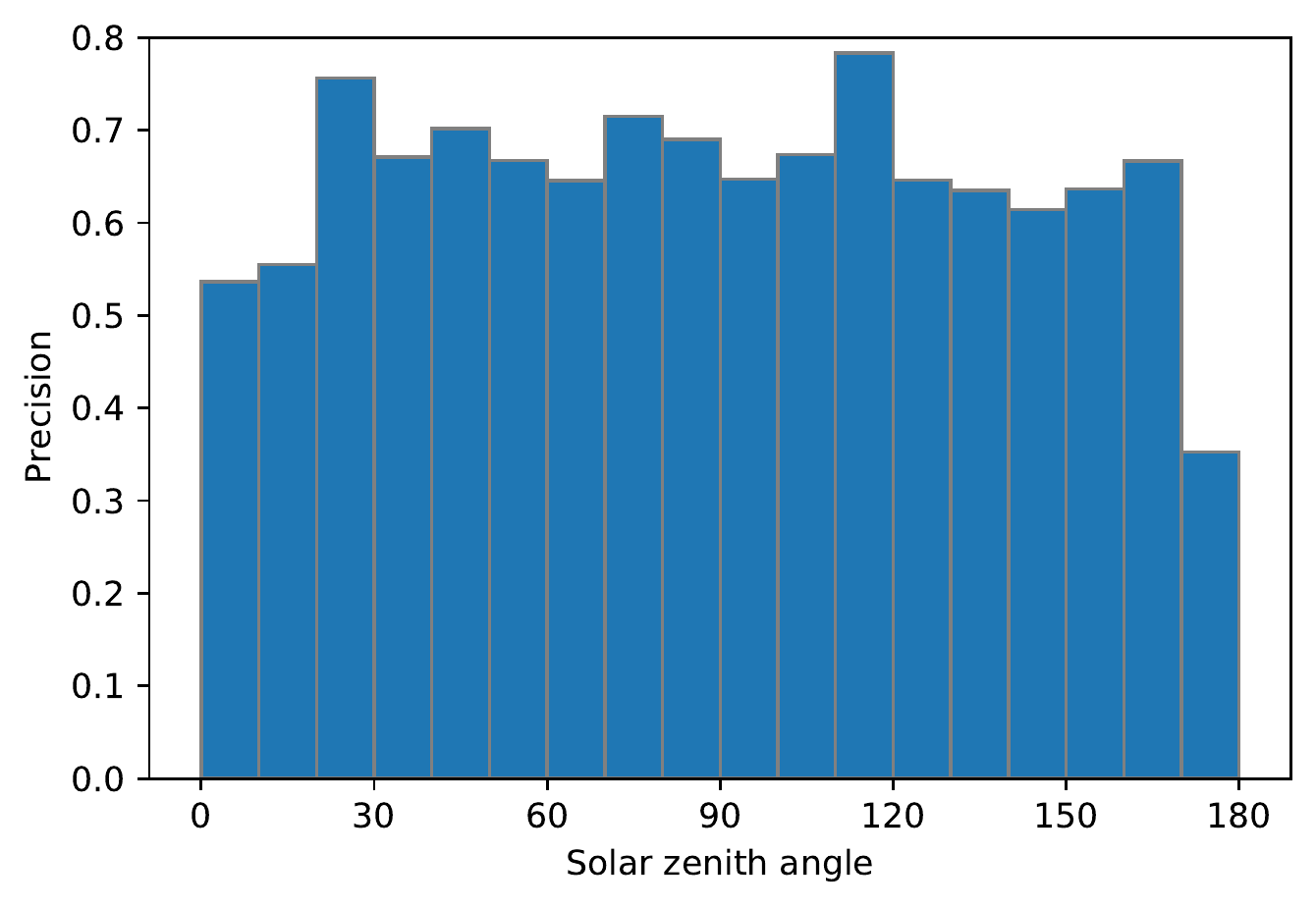}
    \includegraphics[width=.32\textwidth]{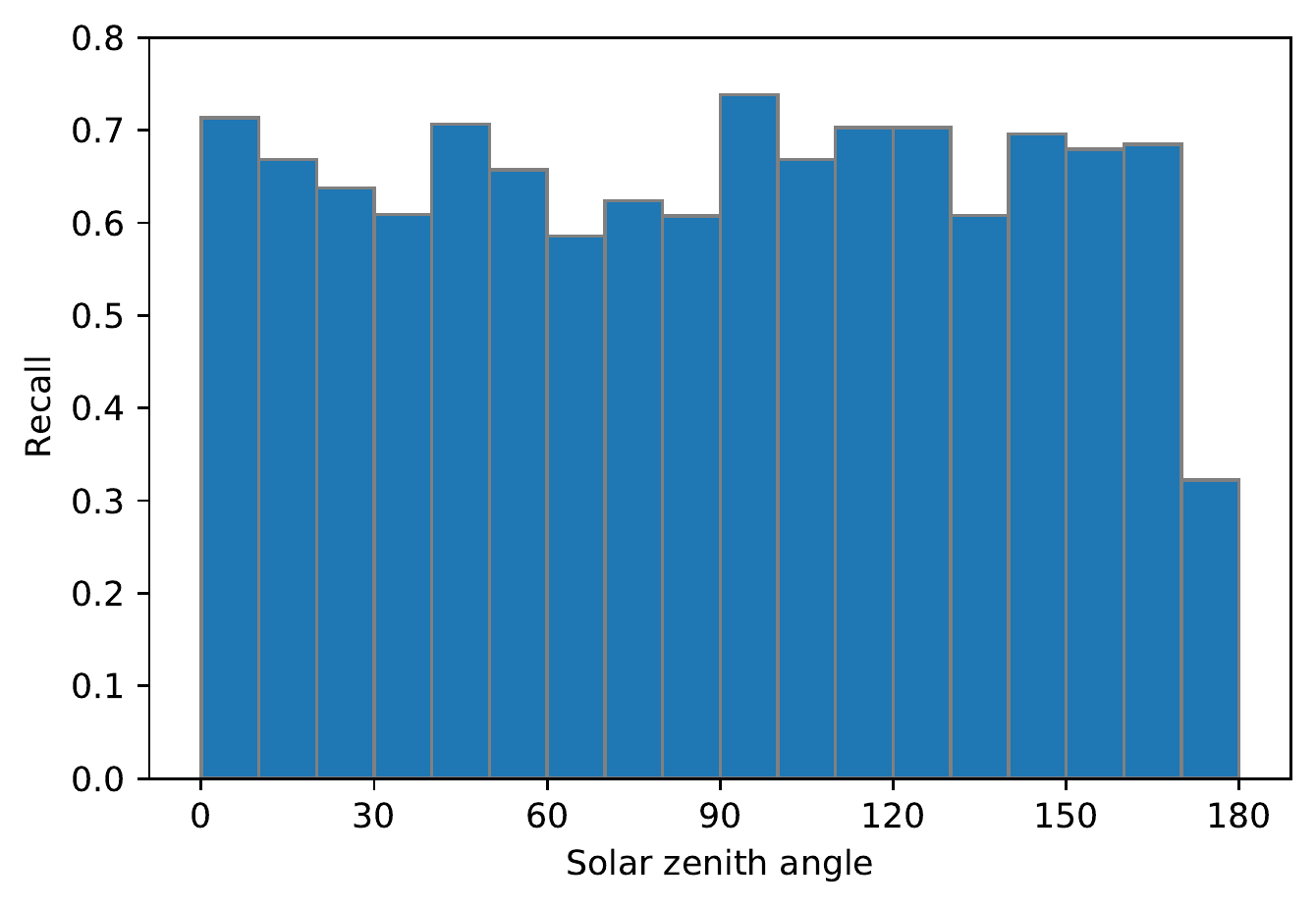}
    \includegraphics[width=.32\textwidth]{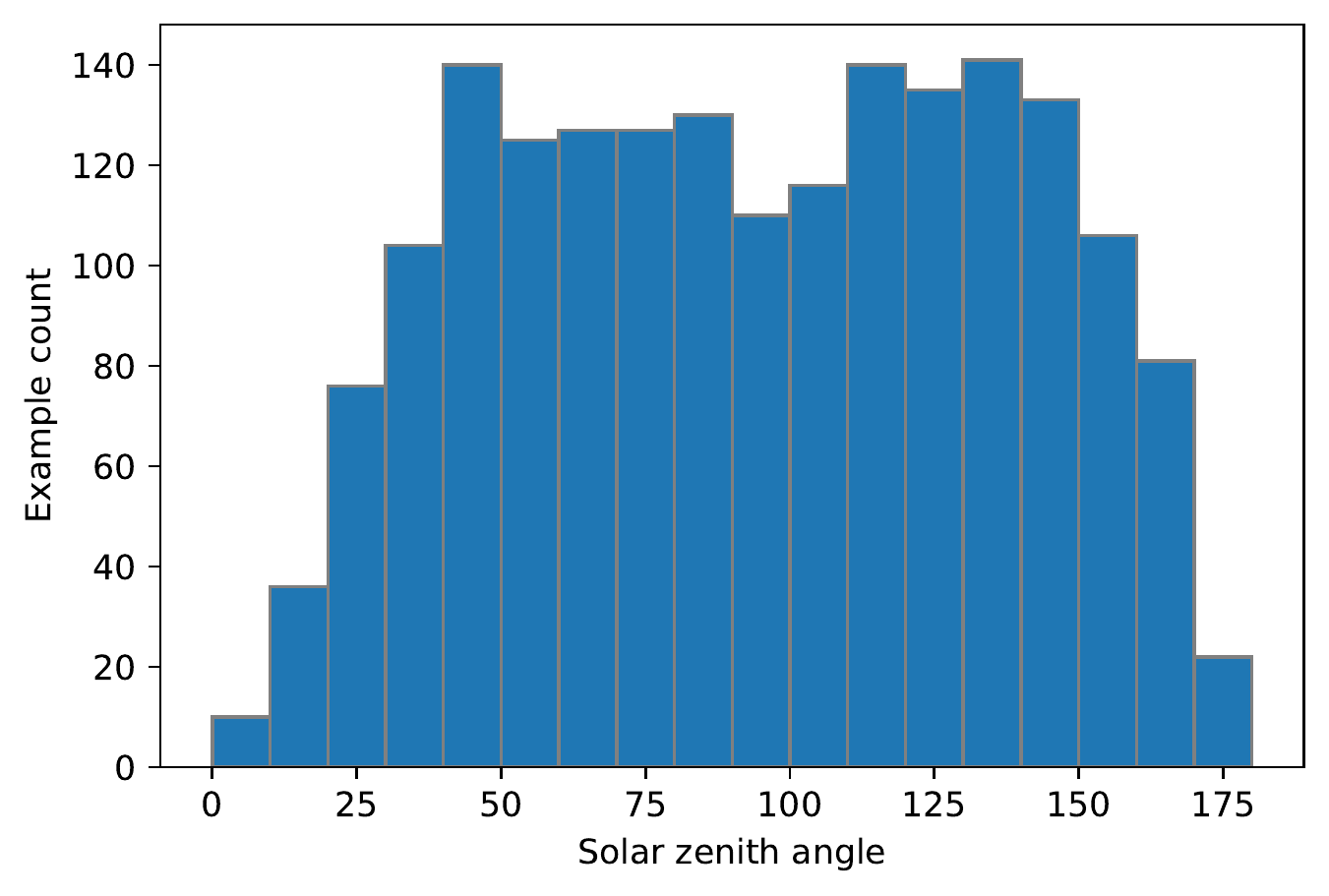}
    \caption{Detector precision (left), recall (middle), and number of examples (right) at different solar zenith angle on the validation set.}
    \label{fig:solar_zenith_performance}
\end{figure*}

\textbf{Performance at different solar zenith angles.}
The solar zenith angle is how many degrees away from "directly overhead" the sun is at any given moment; greater than 90 degrees indicates night scenes. Figure~\ref{fig:solar_zenith_performance} shows the precision and recall of the detector at different solar zenith angles. We observe relatively uniform performance across different solar zenith angles, except in the zenith angle bucket with small number of examples which may lead to noisy results.

\begin{figure*}
    \centering
    \includegraphics[width=.32\textwidth]{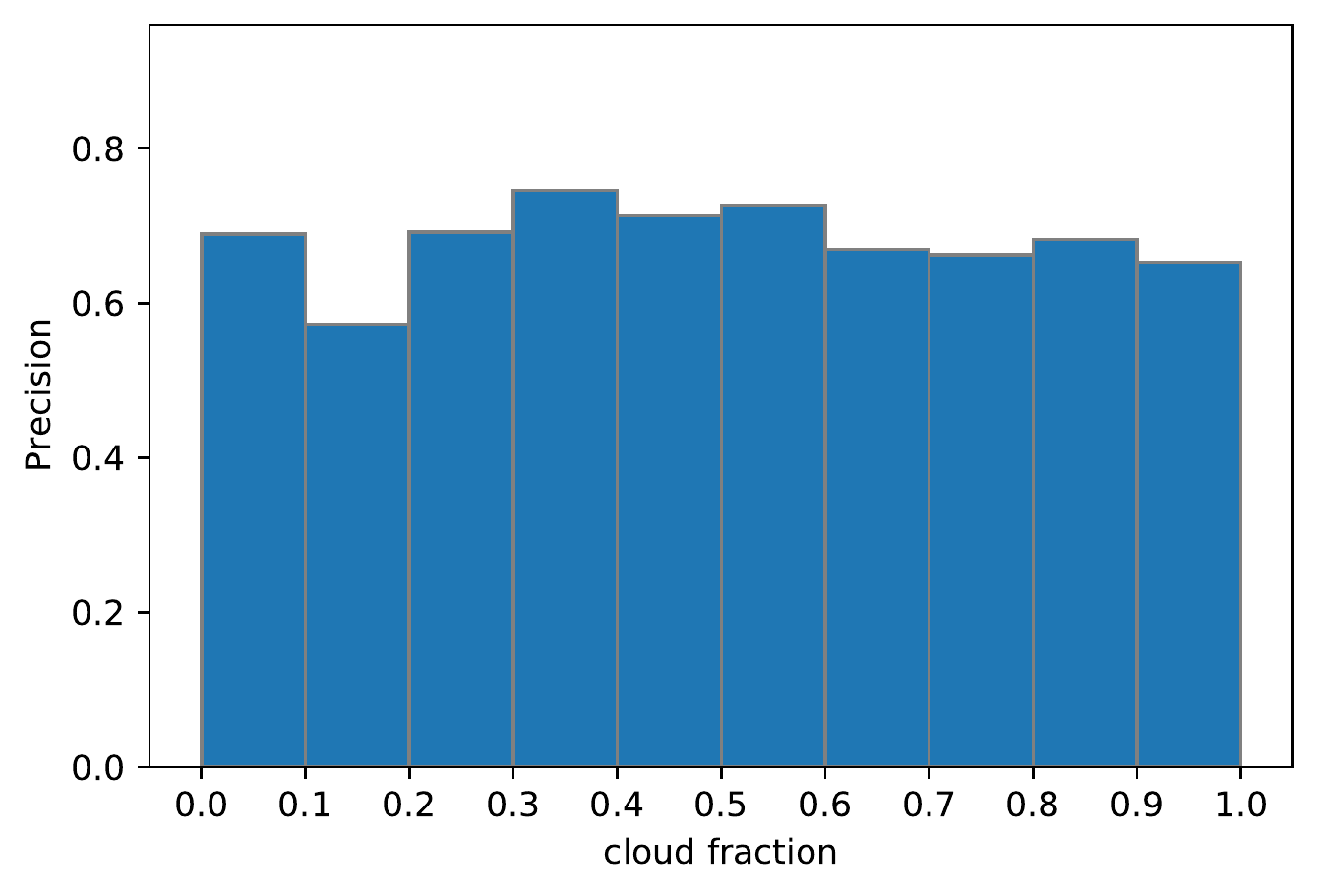}
    \includegraphics[width=.32\textwidth]{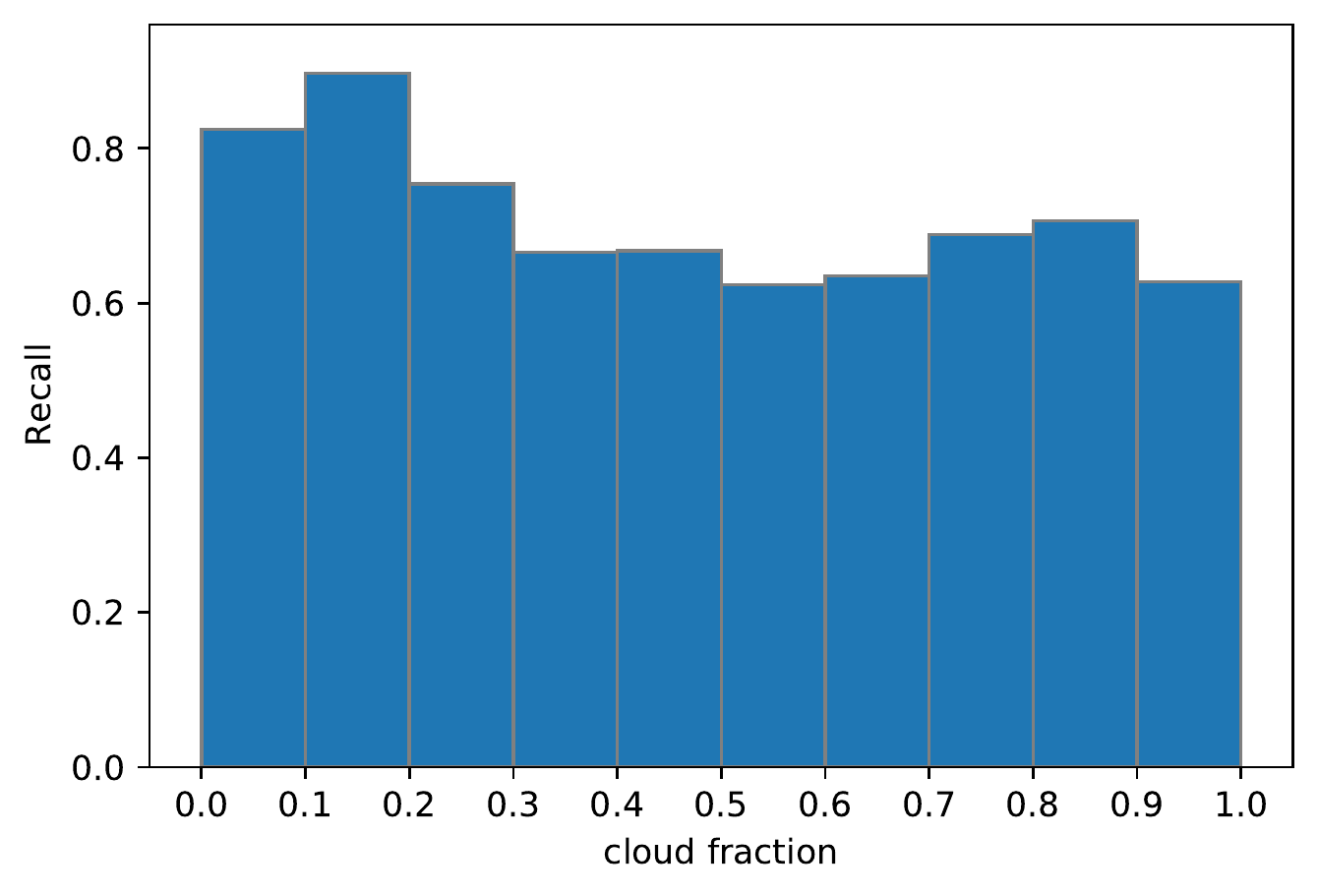}
    \includegraphics[width=.32\textwidth]{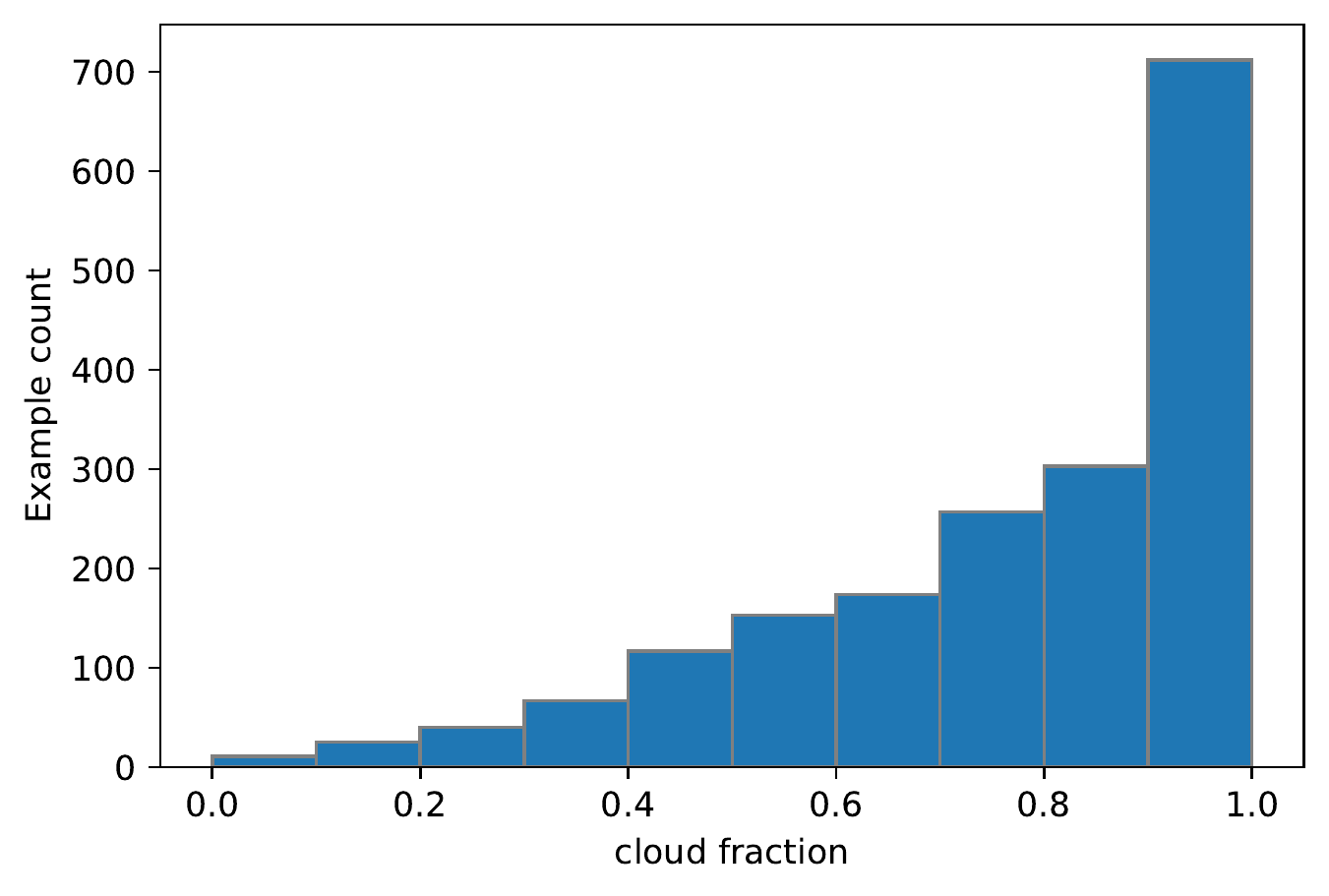}
    \caption{Detector precision (left), recall (middle), and number of examples (right) at different cloud cover fractions on the validation set.}
    \label{fig:cloud_cover_performance}
\end{figure*}

\textbf{Performance at different cloud cover fraction.} 
Contrails in cloudy scenes could be more difficult to recognize. We show the precision and recall (with 0.4 detection threshold) in Figure~\ref{fig:cloud_cover_performance}. The precision is relatively similar across different cloud cover levels, while the recall slightly drops in higher cloud cover levels. 

\textbf{Dataset size.} Larger training datasets often improve the quality of machine learning models. Figure \ref{fig:train_set_size} shows the effect of training with different amounts of training data. More training data improves the detection performance, but diminishing returns are observed at the current scale. This suggests that it may be more important at this point to improve the model architectures, input features, or learning algorithms than it is to collect more labeled data.

\subsection{Comparison with Meijer et al~\cite{meijer2022contrail}}
Meijer et al.~\cite{meijer2022contrail} have also trained a contrail detection model on GOES-16 images. Their model is based on a U-Net architecture with ResNet-18 backbone, and they trained the model on their dataset which consists of 103 labeled images covering the continental US (CONUS) region, with each image having $3000 \times 2000$ pixels. We obtained the evaluation set and detection results from \cite{meijer2022contrail} and here we provide an inter-comparison of both models on both datasets. We report on the images from the year 2018 in the evaluation set from ~\cite{meijer2022contrail}, which contains human labels on the CONUS region for 12 different GOES-16 scans spread throughout the year. We have verified that there is no overlap between the evaluation set and OpenContrail's training set. Since their dataset uses GOES-16 images from 2018 where full-disk images are taken every 15 minutes instead of 10 minutes, we evaluate our single frame model on this dataset instead of the multi-frame model.

Due to a discrepancy in our respective implementations of image projection, during initial comparisons we observed a slight misalignment between the Meijer et al. human labels when compared to our projections of the GOES-16 images. Therefore, we use a ``relaxed'' version of precision and recall as the metric, similar to ~\cite{mnih2010learning}. We consider the relaxed precision represents the fraction of the predicted contrails that are within $\rho$ pixels of the true contrails, and the relaxed recall represents the fraction of true contrails that are within $\rho$ pixels of the predicted contrails. We use $\rho = 2$ in our evaluation.

We compare the detection results without model retraining. We select a threshold value of 0.35 to generate binary detection masks. The results are summarized in Table~\ref{tab:mit_dataset_comparison}. Our model gives better accuracy than~\cite{meijer2022contrail} on our dataset, which is unsurprising given our model is trained on data with a similar distribution. Despite the distribution shift of labels from our training set to the labels from ~\cite{meijer2022contrail}, our model slightly outperforms their model trained on their dataset. This highlights the good quality and the degree of generalization of our contrail detection model trained on our new dataset.

\begin{table}[ht]
\centering
\caption{Model comparison with Meijer et al's detection~\cite{meijer2022contrail} on different datasets. The relaxed precision and recall ($\rho = 2$) of a fixed threshold are reported.}
\label{tab:mit_dataset_comparison}
\begin{tabular}{c|c|c}
      & Relaxed Precision &  Relaxed Recall \\
     \hline
     \multicolumn{3}{c}{Performance on Meijer et al's dataset \cite{meijer2022contrail}} \\
     \hline
     Meijer et al~\cite{meijer2022contrail} & 71.6 & 57.2 \\
     Ours & \textbf{74.5} & \textbf{59.7} \\
     \hline
     \multicolumn{3}{c}{Performance on our dataset} \\
     \hline
     Meijer et al~\cite{meijer2022contrail} & 66.9 & 51.6 \\
     Ours & \textbf{91.3} & \textbf{75.1} \\
     
\end{tabular}
\end{table}

\subsection{Visualization}
\begin{figure*}
    \centering
    \includegraphics[width=.9\linewidth]{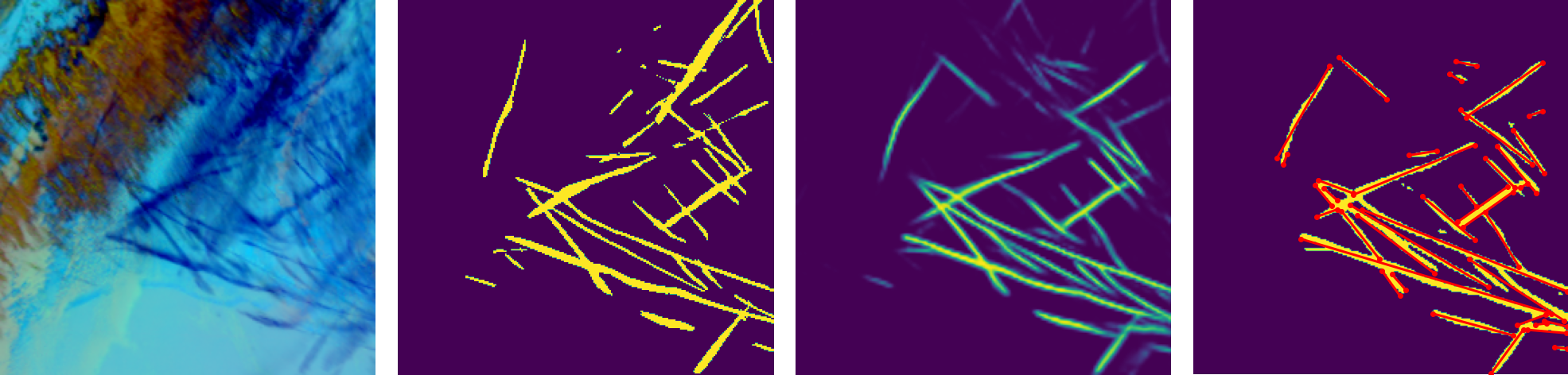}\\
    \vspace{2px}
    \includegraphics[width=.9\linewidth]{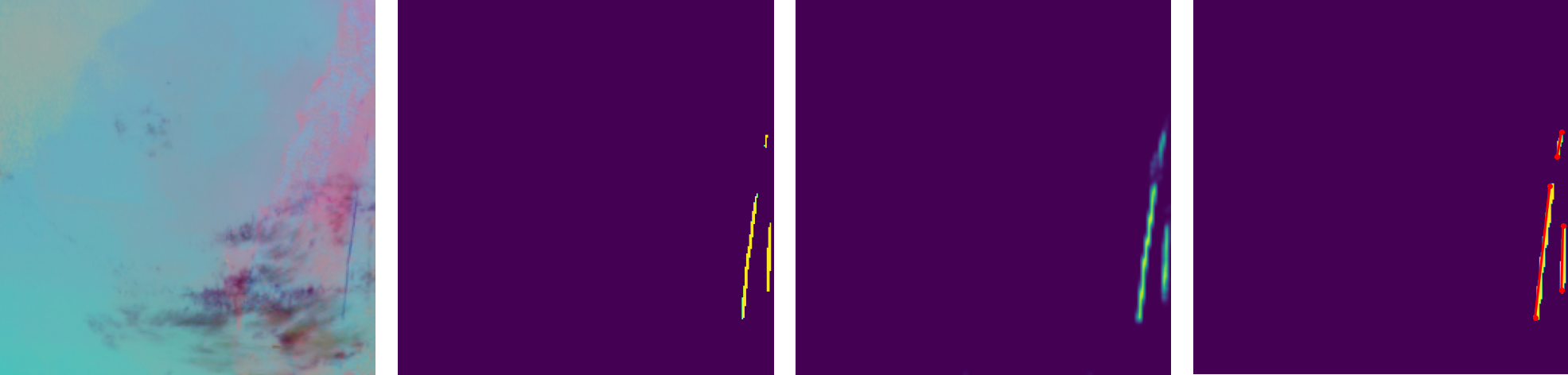}\\
    \vspace{2px}
    \includegraphics[width=.9\linewidth]{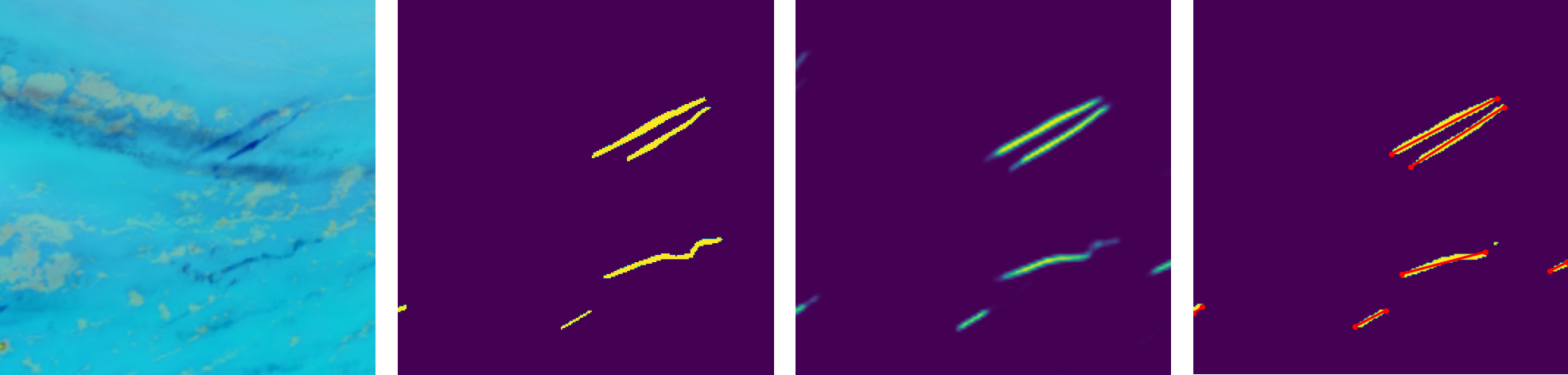}\\
    \vspace{2px}
    \includegraphics[width=.9\linewidth]{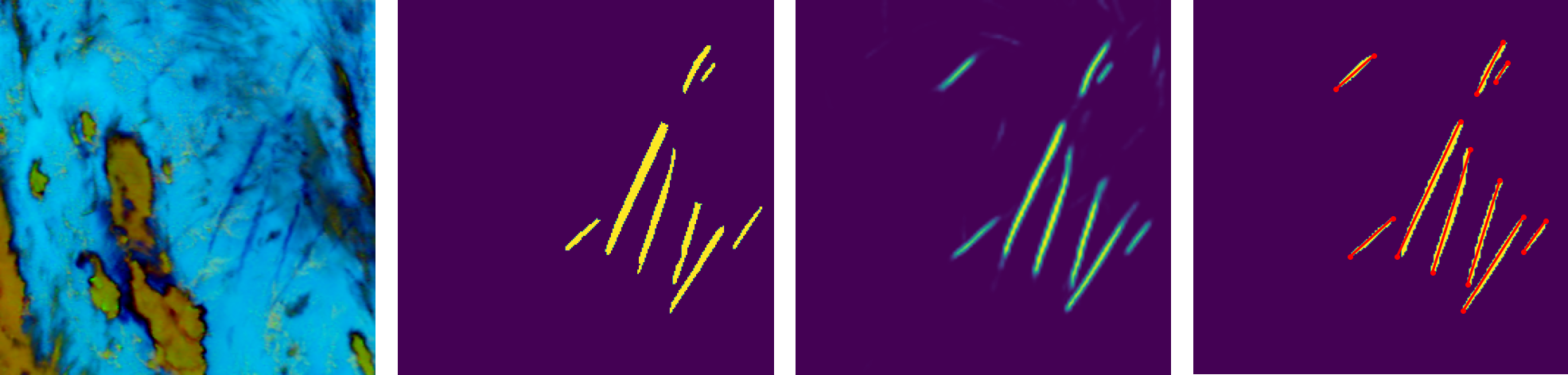}\\
    \vspace{2px}
    \includegraphics[width=.9\linewidth]{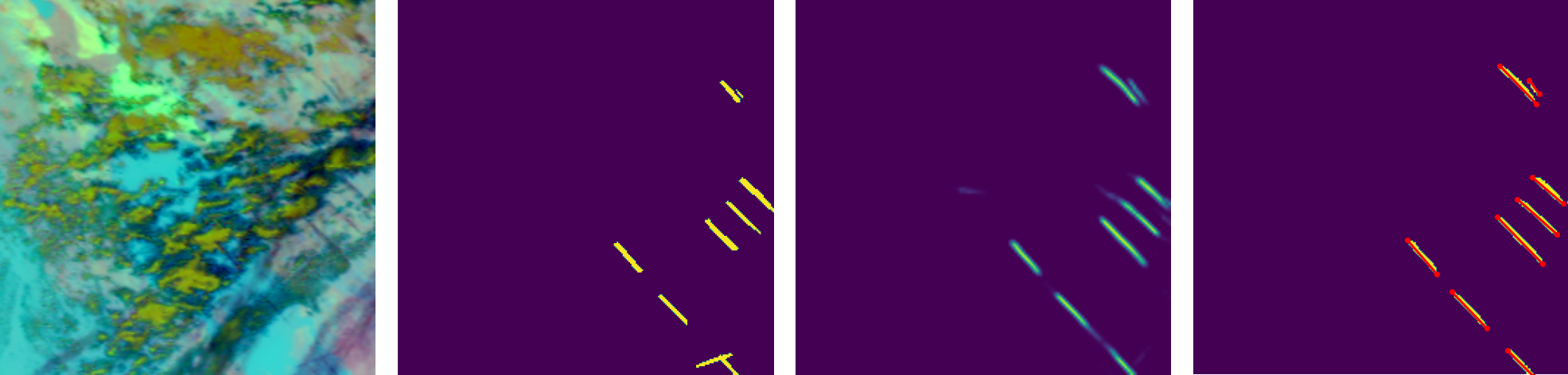}\\
    \caption{Example detection outputs from our model. Columns from left to right: 1) False color images from GOES; 2) Ground truth contrail labels; 3) Pixel outputs from our model; 4) Binarized detection outputs (with threshold 0.35) and the red line segments corresponds to the linearization outputs from the line segment detector.}
    \label{fig:detection}
\end{figure*}

Figure~\ref{fig:detection} shows some qualitative results of our detections. We observe good visual agreement between the model detections and human label masks, and the line segment detector appears able to convert the detection mask to linear segments reasonably well. The model sometimes misses some detections when there are many contrails in the scene. It is also clear the model generally does not assign much (if any) probability to anthropogenic contrail-cirrus clouds that are adjacent to still-linear contrails. This is unsurprising given the instructions we have provided to labelers (to only mark linear contrails), but it has an important implication for using the detection model as the basis of contrail warming assessments: we expect the warming from contrail pixels detected by the model to comprise a lower-bound on the actual contrail warming impact.

\section{Contrail Coverage}
\begin{figure*}
    \includegraphics[height=270px]{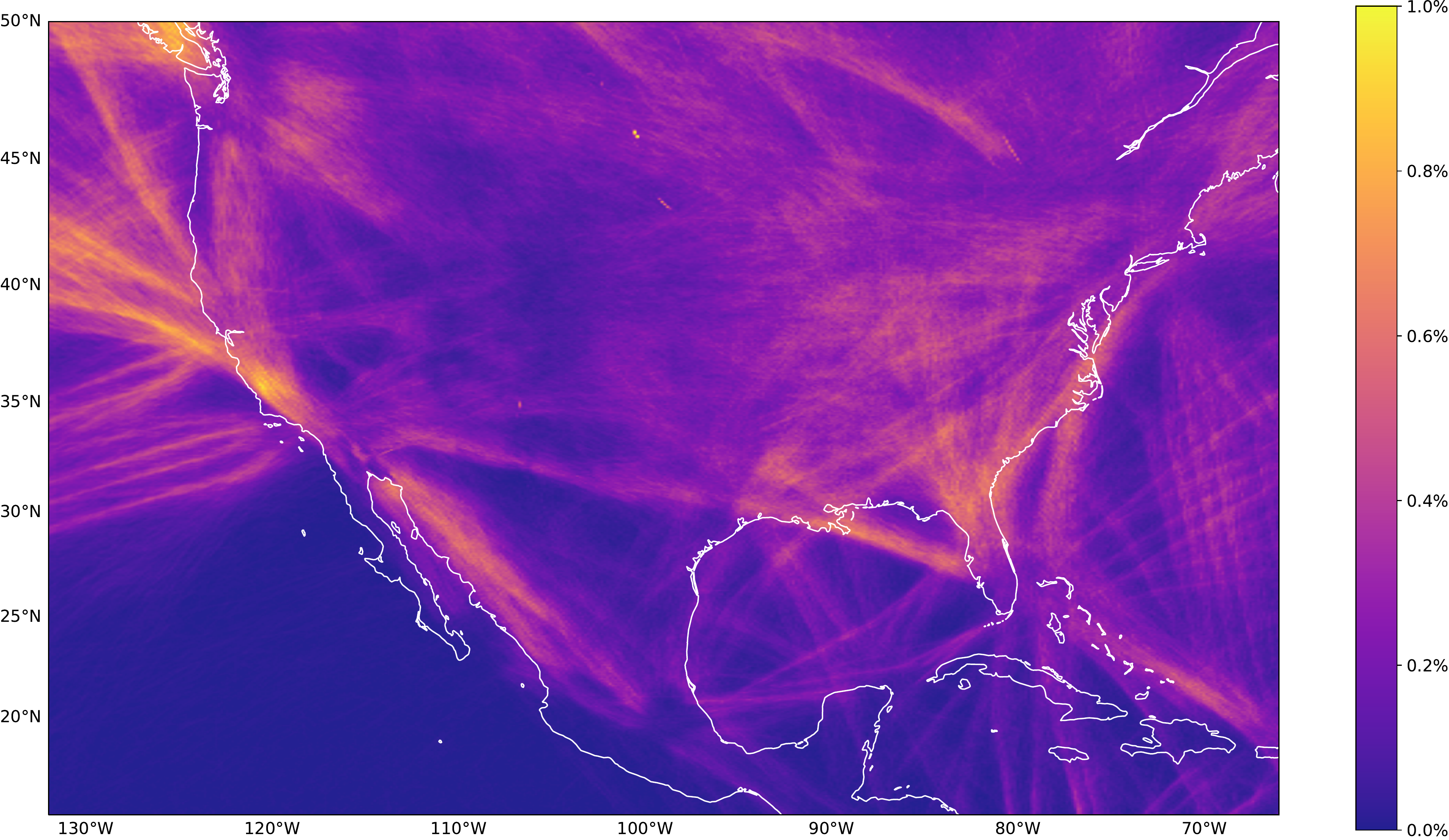} \\
    \vspace{4px}\\
    \includegraphics[height=270px]{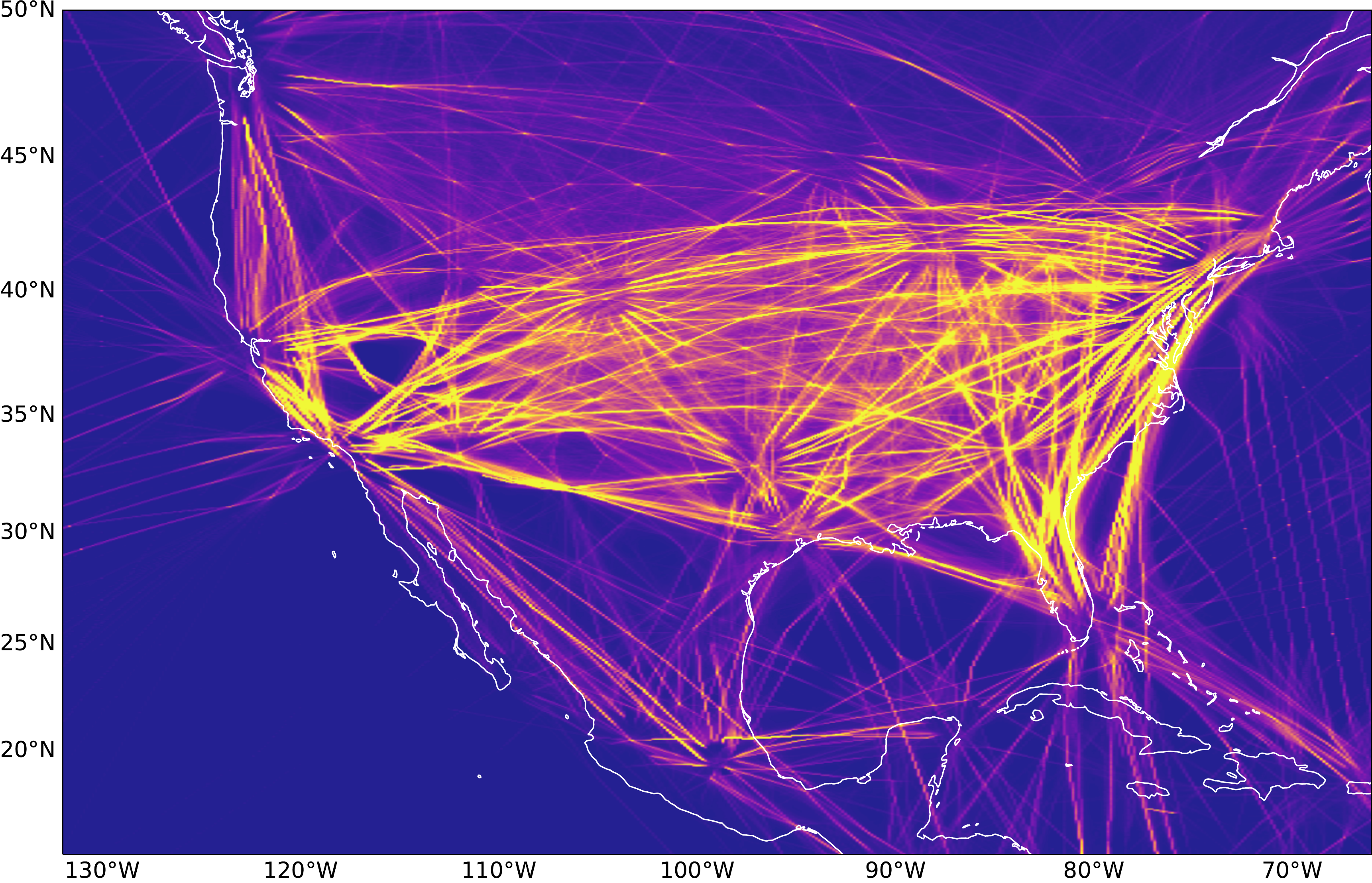}
    \caption{Top: averaged detected contrail mask, 2018 -- 2019. Bottom: flight waypoint density, 2018 -- 2019.}
    \label{fig:averaged_mask}
\end{figure*}

To compute contrail coverage visuals and statistics we run our contrail detection model as an overlapping sliding window, and then stitch the single-frame model detection outputs for the CONUS region in year 2018 -- 2019. To give an unbiased contrail coverage estimate, following Meijer et al. \cite{meijer2022contrail} we select a per-pixel detection threshold (for our model, 0.4) which gives a precision rate and recall rate on our validation set that are approximately equal. We show the average contrail coverage in 2018 -- 2019 in the CONUS region and the corresponding flight density (computed from FlightAware data) in Figure \ref{fig:averaged_mask}. The contrails cluster along the high traffic flight routes, which confirms the pattern from \cite{meijer2022contrail}. The average linear contrail coverage in the considered domain is 0.19\% which is similar to 0.17\% reported in~\cite{meijer2022contrail}.

\begin{figure}
    \centering
    \includegraphics[width=.99\linewidth]{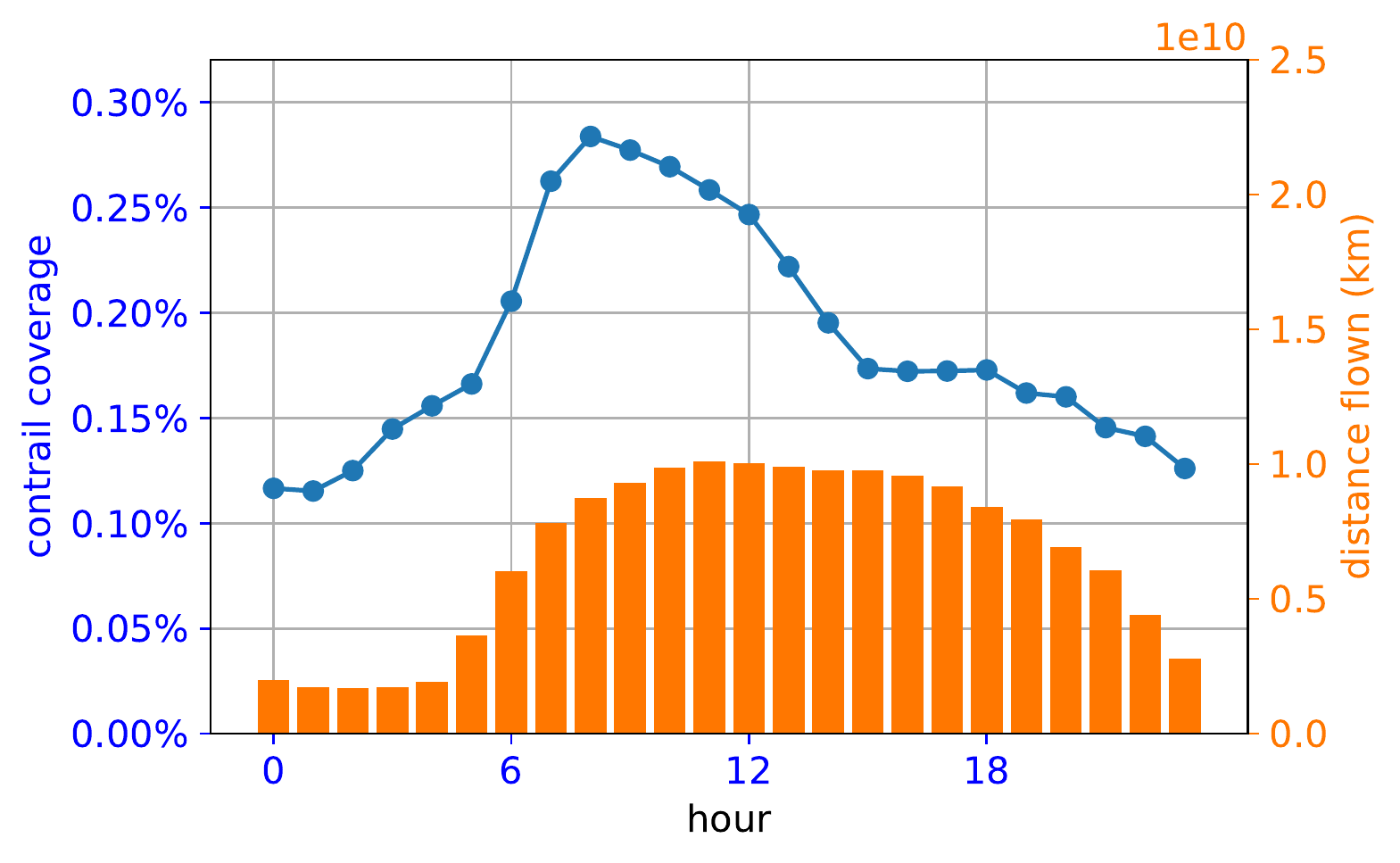}
    \caption{Linear contrail coverage and flight density by time of day. The blue curve shows the linear contrail coverage derived from our contrail detector and the orange bar shows the flight densities.}
    \label{fig:coverage_vs_flight}
\end{figure}

Figure~\ref{fig:coverage_vs_flight} shows our estimated linear contrail coverage and flight density by the time of day in 2019. The local hour of the contrails are approximated by $\mathrm{hour}_{\mathrm{UTC}} + \mathrm{longitude} / 360 \times 24$. We observe a similar diurnal pattern as in previous work~\cite{meijer2022contrail}: the contrail coverage is lowest around midnight with about 0.1\% coverage. It peaks at around 8am with about 0.21\% contrail coverage (as the  distance flown goes up) but then drops in the afternoon despite the flight distance remaining high. 

\begin{figure*}
    \centering
    \includegraphics[width=.9\linewidth]{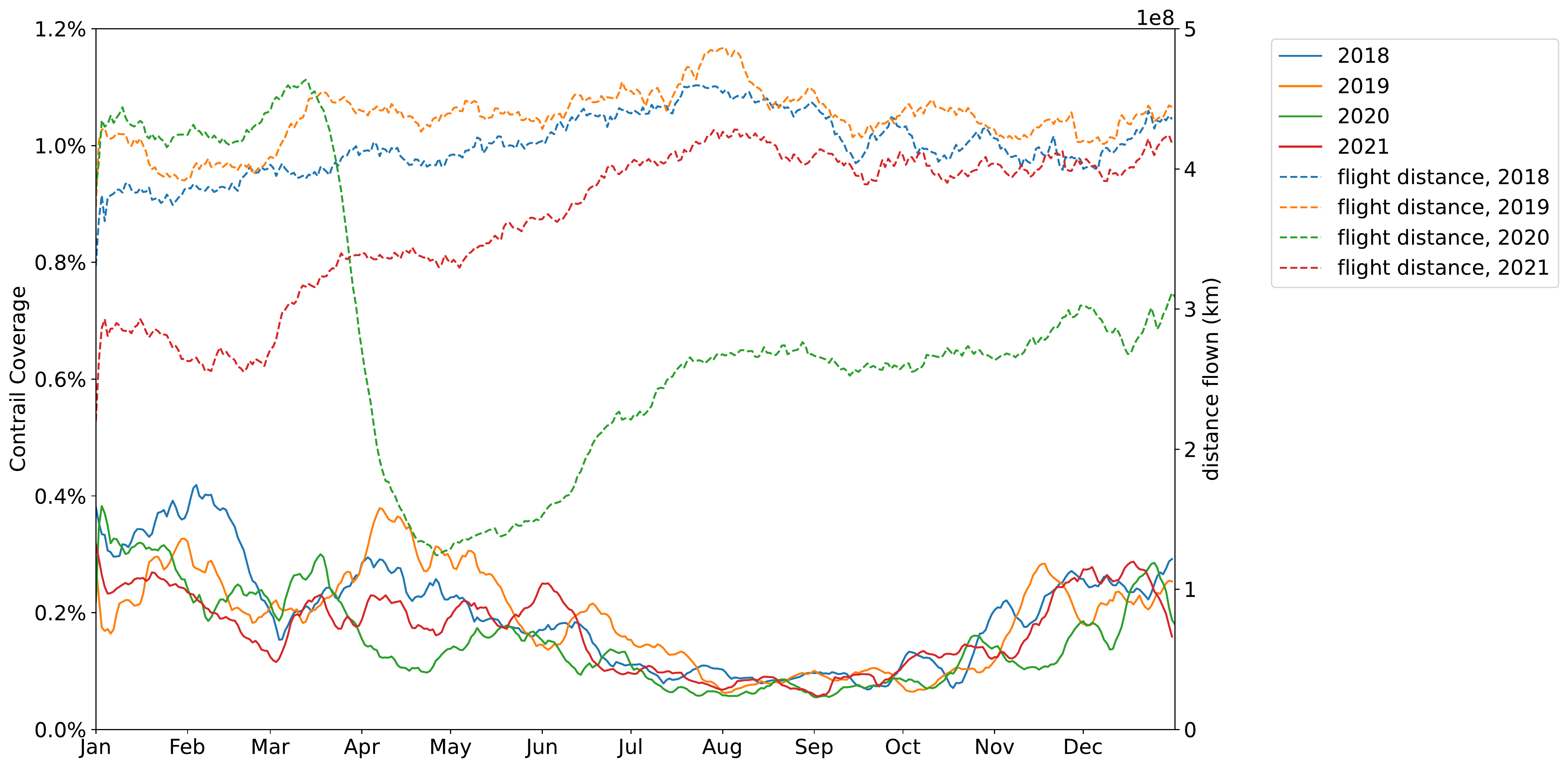}
    \caption{Linear contrail coverage from our single-frame model with flight distance by day of year from 2018 to 2021. 15-day rolling means are displayed. }
    \label{fig:coverage_day_of_year}
\end{figure*}

Figure~\ref{fig:coverage_day_of_year} shows the contrail coverage vs flight distance by the day of year from 2018-2021. We observe similar trends to those observed by Meijer et al~\cite{meijer2022contrail}. Contrail coverage drops as the flight distance drops significantly in April 2020, due to COVID-19. We also observe a seasonal effect of contrail coverage, where fewer contrails are observed consistently in the summer months across multiple years, regardless of the flight distance flown.

\section{Discussion}
While our dataset provides high quality contrail labels on GOES-16 satellite images, some contrails may not be observable from geostationary satellites and thus cannot be labeled by human raters. For example, young contrails may already be deformed (or dissipated) before they are visible from geostationary orbit. Also, contrails forming below existing optically thick cirrus are unlikely to be recognizable by human labelers; however it has been suggested that those contrails do not generate a significant amount of warming~\cite{teoh2022aviation}. Further research is needed to empirically determine the fraction of contrails and their corresponding warming which can be detected in geoestationary satellite images.

Our current system detects contrails at each target frame independently, and does not provide correspondence of contrails between difference frames. Such correspondence could be useful for identifying the contrail age, and improving flight attribution (i.e., which flight caused the contrail). It is also currently estimated that a large fraction of contrail warming happens after contrails are old enough they are likely no longer linear ~\cite{teoh2022aviation}. Our proposed models and labels only include linear-shaped contrails, and do not identify deformed contrail-cirrus clouds which are no longer distinguishable from natural cirrus without tracing back to when they are formed. To fully assess the warming impact of contrails, it may be crucial to track contrail pixels for an extended period of time once they are detected. Such a contrail tracking system can be built based on the linear contrail models demonstrated in this work.

Due to the high cost of pixel-wise labeling of contrails, our proposed dataset is relatively small compared to large scale natural image datasets like ImageNet (20k vs 1M examples). Large scale labeled datasets for geostationary satellite images are unfortunately not currently available for model pre-training. Therefore, our proposed models are initialized on checkpoints trained on natural image or video datasets such as ImageNet or Kinetics, and their benefits may be limited by domain mismatch between internet images and satellite images. 
Self-supervised and semi-supervised machine learning techniques that leverage unlabeled data have been shown to improve machine learning model performance for remote sensing segmentation tasks~\cite{patel2021evaluating}, and should be explored here in the future. %

Our work here can be extended to cover other geostationary satellites. We plan to collect labels on other geostationary satellites (e.g. Himawari-8~\cite{bessho2016introduction} and Meteosat-11~\cite{schmetz2002introduction}) to cover other regions including Europe and Asia-Pacific. We believe that models trained on the proposed OpenContrails dataset based on GOES-16 can be used for transfer learning to obtain high quality contrail detectors on other satellites with fewer human labels. 

\section{Conclusion}
We present here a large human-labeled contrail detection benchmark dataset based on GOES-16 ABI imagery.
Using it, we explore several contrail detection model architectures, and achieve a new state of the art in linear contrail detection on GOES-16 ABI imagery. We show that leveraging temporal context improves contrail detection, and confirm previously reported seasonal and diurnal contrail coverage trends. The contrail detection dataset and the detection model outputs are made publicly available. We believe these contributions will advance contrail empirical studies including contrail detection, climate impact assessment, and mitigation.

\section{Acknowledgement}
The authors gratefully acknowledge: Rachel Soh, Jessica Ferguson, Jeanie Pearson, Ladislav Honsa and Rob von Behren for assistance acquiring and processing flight path data; Nathan Kiner for assistance setting up the contrail annotation pipeline; and the Google Research data operations team for labeling the GOES-16 imagery.

{\small
\bibliographystyle{IEEEtran}
\bibliography{ref}

\begin{thebibliography}{10}
\providecommand{\url}[1]{#1}
\csname url@samestyle\endcsname
\providecommand{\newblock}{\relax}
\providecommand{\bibinfo}[2]{#2}
\providecommand{\BIBentrySTDinterwordspacing}{\spaceskip=0pt\relax}
\providecommand{\BIBentryALTinterwordstretchfactor}{4}
\providecommand{\BIBentryALTinterwordspacing}{\spaceskip=\fontdimen2\font plus
\BIBentryALTinterwordstretchfactor\fontdimen3\font minus
  \fontdimen4\font\relax}
\providecommand{\BIBforeignlanguage}[2]{{%
\expandafter\ifx\csname l@#1\endcsname\relax
\typeout{** WARNING: IEEEtran.bst: No hyphenation pattern has been}%
\typeout{** loaded for the language `#1'. Using the pattern for}%
\typeout{** the default language instead.}%
\else
\language=\csname l@#1\endcsname
\fi
#2}}
\providecommand{\BIBdecl}{\relax}
\BIBdecl

\bibitem{lee2021contribution}
D.~S. Lee, D.~Fahey, A.~Skowron, M.~Allen, U.~Burkhardt, Q.~Chen, S.~Doherty,
  S.~Freeman, P.~Forster, J.~Fuglestvedt \emph{et~al.}, ``The contribution of
  global aviation to anthropogenic climate forcing for 2000 to 2018,''
  \emph{Atmospheric Environment}, vol. 244, p. 117834, 2021.

\bibitem{bickel2020estimating}
M.~Bickel, M.~Ponater, L.~Bock, U.~Burkhardt, and S.~Reineke, ``Estimating the
  effective radiative forcing of contrail cirrus,'' \emph{Journal of Climate},
  vol.~33, no.~5, pp. 1991--2005, 2020.

\bibitem{avila2019reducing}
D.~Avila, L.~Sherry, and T.~Thompson, ``Reducing global warming by airline
  contrail avoidance: A case study of annual benefits for the contiguous united
  states,'' \emph{Transportation Research Interdisciplinary Perspectives},
  vol.~2, p. 100033, 2019.

\bibitem{teoh2020mitigating}
R.~Teoh, U.~Schumann, A.~Majumdar, and M.~E. Stettler, ``Mitigating the climate
  forcing of aircraft contrails by small-scale diversions and technology
  adoption,'' \emph{Environmental Science \& Technology}, vol.~54, no.~5, pp.
  2941--2950, 2020.

\bibitem{gierens2020}
\BIBentryALTinterwordspacing
K.~Gierens, S.~Matthes, and S.~Rohs, ``How well can persistent contrails be
  predicted?'' \emph{Aerospace}, vol.~7, no.~12, 2020. [Online]. Available:
  \url{https://www.mdpi.com/2226-4310/7/12/169}
\BIBentrySTDinterwordspacing

\bibitem{deng2009imagenet}
J.~Deng, W.~Dong, R.~Socher, L.-J. Li, K.~Li, and L.~Fei-Fei, ``Imagenet: A
  large-scale hierarchical image database,'' in \emph{Proceedings of IEEE
  conference on Computer Vision and Pattern Recognition}, 2009, pp. 248--255.

\bibitem{mnist}
L.~Deng, ``The mnist database of handwritten digit images for machine learning
  research,'' \emph{IEEE Signal Processing Magazine}, no.~6, pp. 141--142,
  2012.

\bibitem{meijer2022contrail}
V.~R. Meijer, L.~Kulik, S.~D. Eastham, F.~Allroggen, R.~L. Speth, S.~Karaman,
  and S.~R. Barrett, ``Contrail coverage over the united states before and
  during the covid-19 pandemic,'' \emph{Environmental Research Letters},
  vol.~17, no.~3, p. 034039, 2022.

\bibitem{zhang2018contrail}
G.~Zhang, J.~Zhang, and J.~Shang, ``Contrail recognition with convolutional
  neural network and contrail parameterizations evaluation,'' \emph{SOLA},
  vol.~14, pp. 132--137, 2018.

\bibitem{chen2013simulated}
C.-C. Chen and A.~Gettelman, ``Simulated radiative forcing from contrails and
  contrail cirrus,'' \emph{Atmospheric Chemistry and Physics}, vol.~13, no.~24,
  pp. 12\,525--12\,536, 2013.

\bibitem{schumann2012contrail}
U.~Schumann, ``A contrail cirrus prediction model,'' \emph{Geoscientific Model
  Development}, vol.~5, no.~3, pp. 543--580, 2012.

\bibitem{mannstein1999operational}
H.~Mannstein, R.~Meyer, and P.~Wendling, ``Operational detection of contrails
  from noaa-avhrr-data,'' \emph{International Journal of Remote Sensing},
  vol.~20, no.~8, pp. 1641--1660, 1999.

\bibitem{meyer2002regional}
R.~Meyer, H.~Mannstein, R.~Meerk{\"o}tter, U.~Schumann, and P.~Wendling,
  ``Regional radiative forcing by line-shaped contrails derived from satellite
  data,'' \emph{Journal of Geophysical Research: Atmospheres}, vol. 107, no.
  D10, pp. ACL--17, 2002.

\bibitem{vazquez2015contrail}
M.~V{\'a}zquez-Navarro, H.~Mannstein, and S.~Kox, ``Contrail life cycle and
  properties from 1 year of msg/seviri rapid-scan images,'' \emph{Atmospheric
  Chemistry and Physics}, vol.~15, no.~15, pp. 8739--8749, 2015.

\bibitem{mccloskey2021human}
K.~McCloskey, S.~Geraedts, B.~Jackman, V.~R. Meijer, E.~Brand, D.~Fork, J.~C.
  Platt, C.~Elkin, and C.~Van~Arsdale, ``A human-labeled landsat-8 contrails
  dataset,'' in \emph{International Conference on Machine Learning workshop on
  Tackling Climate Change with Machine Learning}, 2021.

\bibitem{schmit2017closer}
T.~J. Schmit, P.~Griffith, M.~M. Gunshor, J.~M. Daniels, S.~J. Goodman, and
  W.~J. Lebair, ``A closer look at the abi on the goes-r series,''
  \emph{Bulletin of the American Meteorological Society}, vol.~98, no.~4, pp.
  681--698, 2017.

\bibitem{ma2019deep}
L.~Ma, Y.~Liu, X.~Zhang, Y.~Ye, G.~Yin, and B.~A. Johnson, ``Deep learning in
  remote sensing applications: A meta-analysis and review,'' \emph{ISPRS
  journal of photogrammetry and remote sensing}, vol. 152, pp. 166--177, 2019.

\bibitem{zhu2017deep}
X.~X. Zhu, D.~Tuia, L.~Mou, G.-S. Xia, L.~Zhang, F.~Xu, and F.~Fraundorfer,
  ``Deep learning in remote sensing: A comprehensive review and list of
  resources,'' \emph{IEEE Geoscience and Remote Sensing Magazine}, vol.~5,
  no.~4, pp. 8--36, 2017.

\bibitem{he2016deep}
K.~He, X.~Zhang, S.~Ren, and J.~Sun, ``Deep residual learning for image
  recognition,'' in \emph{Proceedings of the IEEE conference on Computer Vision
  and Pattern Recognition}, 2016, pp. 770--778.

\bibitem{chen2017rethinking}
L.-C. Chen, G.~Papandreou, F.~Schroff, and H.~Adam, ``Rethinking atrous
  convolution for semantic image segmentation,'' \emph{arXiv preprint
  arXiv:1706.05587}, 2017.

\bibitem{chen2018encoder}
L.-C. Chen, Y.~Zhu, G.~Papandreou, F.~Schroff, and H.~Adam, ``Encoder-decoder
  with atrous separable convolution for semantic image segmentation,'' in
  \emph{Proceedings of the European conference on computer vision (ECCV)},
  2018, pp. 801--818.

\bibitem{definition2019users}
\BIBentryALTinterwordspacing
GOES-R, ``Goes-r series product definition and users' guide,'' 2019. [Online].
  Available: \url{https://www.goes-r.gov/products/docs/PUG-L2+-vol5.pdf}
\BIBentrySTDinterwordspacing

\bibitem{hersbach2020era5}
H.~Hersbach, B.~Bell, P.~Berrisford, S.~Hirahara, A.~Hor{\'a}nyi,
  J.~Mu{\~n}oz-Sabater, J.~Nicolas, C.~Peubey, R.~Radu, D.~Schepers
  \emph{et~al.}, ``The era5 global reanalysis,'' \emph{Quarterly Journal of the
  Royal Meteorological Society}, vol. 146, no. 730, pp. 1999--2049, 2020.

\bibitem{bogacki19893}
P.~Bogacki and L.~F. Shampine, ``A 3 (2) pair of runge-kutta formulas,''
  \emph{Applied Mathematics Letters}, vol.~2, no.~4, pp. 321--325, 1989.

\bibitem{gierensbook}
K.~Gierens, P.~Spichtinger, and U.~Schumann, \emph{Atmospheric Physics:
  Background - Methods - Trends}.\hskip 1em plus 0.5em minus 0.4em\relax
  Springer, Heidelberg, 2012.

\bibitem{juan2019graph}
D.-C. Juan, C.-T. Lu, Z.~Li, F.~Peng, A.~Timofeev, Y.-T. Chen, Y.~Gao,
  T.~Duerig, A.~Tomkins, and S.~Ravi, ``Graph-rise: Graph-regularized image
  semantic embedding,'' \emph{arXiv preprint arXiv:1902.10814}, 2019.

\bibitem{kulik2019satellite}
L.~Kulik, ``Satellite-based detection of contrails using deep learning,'' Ph.D.
  dissertation, Massachusetts Institute of Technology, 2019.

\bibitem{bello2021revisiting}
I.~Bello, W.~Fedus, X.~Du, E.~D. Cubuk, A.~Srinivas, T.-Y. Lin, J.~Shlens, and
  B.~Zoph, ``Revisiting resnets: Improved training and scaling strategies,''
  \emph{Advances in Neural Information Processing Systems}, vol.~34, pp.
  22\,614--22\,627, 2021.

\bibitem{he2019bag}
T.~He, Z.~Zhang, H.~Zhang, Z.~Zhang, J.~Xie, and M.~Li, ``Bag of tricks for
  image classification with convolutional neural networks,'' in
  \emph{Proceedings of the IEEE/CVF Conference on Computer Vision and Pattern
  Recognition}, 2019, pp. 558--567.

\bibitem{hu2018squeeze}
J.~Hu, L.~Shen, and G.~Sun, ``Squeeze-and-excitation networks,'' in
  \emph{Proceedings of the IEEE conference on Computer Vision and Pattern
  Recognition}, 2018, pp. 7132--7141.

\bibitem{huang2016deep}
G.~Huang, Y.~Sun, Z.~Liu, D.~Sedra, and K.~Q. Weinberger, ``Deep networks with
  stochastic depth,'' in \emph{Proceedings of the European conference on
  computer vision (ECCV)}.\hskip 1em plus 0.5em minus 0.4em\relax Springer,
  2016, pp. 646--661.

\bibitem{wang2018non}
X.~Wang, R.~Girshick, A.~Gupta, and K.~He, ``Non-local neural networks,'' in
  \emph{Proceedings of the IEEE conference on Computer Vision and Pattern
  Recognition}, 2018, pp. 7794--7803.

\bibitem{qiu2017learning}
Z.~Qiu, T.~Yao, and T.~Mei, ``Learning spatio-temporal representation with
  pseudo-3d residual networks,'' in \emph{Proceedings of the IEEE International
  Conference on Computer Vision}, 2017, pp. 5533--5541.

\bibitem{kay2017kinetics}
W.~Kay, J.~Carreira, K.~Simonyan, B.~Zhang, C.~Hillier, S.~Vijayanarasimhan,
  F.~Viola, T.~Green, T.~Back, P.~Natsev \emph{et~al.}, ``The kinetics human
  action video dataset,'' \emph{arXiv preprint arXiv:1705.06950}, 2017.

\bibitem{von2012lsd}
R.~G. Von~Gioi, J.~Jakubowicz, J.-M. Morel, and G.~Randall, ``Lsd: A line
  segment detector,'' \emph{Image Processing On Line}, vol.~2, pp. 35--55,
  2012.

\bibitem{tan2019efficientnet}
M.~Tan and Q.~Le, ``Efficientnet: Rethinking model scaling for convolutional
  neural networks,'' in \emph{International Conference on Machine
  Learning}.\hskip 1em plus 0.5em minus 0.4em\relax PMLR, 2019, pp. 6105--6114.

\bibitem{lin2017focal}
T.-Y. Lin, P.~Goyal, R.~Girshick, K.~He, and P.~Doll{\'a}r, ``Focal loss for
  dense object detection,'' in \emph{Proceedings of the IEEE International
  Conference on Computer Vision}, 2017, pp. 2980--2988.

\bibitem{mnih2010learning}
V.~Mnih and G.~E. Hinton, ``Learning to detect roads in high-resolution aerial
  images,'' in \emph{Proceedings of the European conference on computer vision
  (ECCV)}.\hskip 1em plus 0.5em minus 0.4em\relax Springer, 2010, pp. 210--223.

\bibitem{teoh2022aviation}
R.~Teoh, U.~Schumann, E.~Gryspeerdt, M.~Shapiro, J.~Molloy, G.~Koudis,
  C.~Voigt, and M.~Stettler, ``Aviation contrail climate effects in the north
  atlantic from 2016--2021,'' \emph{Atmospheric Chemistry and Physics
  Discussions}, pp. 1--27, 2022.

\bibitem{patel2021evaluating}
C.~Patel, S.~Sharma, and V.~Gulshan, ``Evaluating self and semi-supervised
  methods for remote sensing segmentation tasks,'' \emph{arXiv preprint
  arXiv:2111.10079}, 2021.

\bibitem{bessho2016introduction}
K.~Bessho, K.~Date, M.~Hayashi, A.~Ikeda, T.~Imai, H.~Inoue, Y.~Kumagai,
  T.~Miyakawa, H.~Murata, T.~Ohno \emph{et~al.}, ``An introduction to
  himawari-8/9—japan’s new-generation geostationary meteorological
  satellites,'' \emph{Journal of the Meteorological Society of Japan. Ser. II},
  vol.~94, no.~2, pp. 151--183, 2016.

\bibitem{schmetz2002introduction}
J.~Schmetz, P.~Pili, S.~Tjemkes, D.~Just, J.~Kerkmann, S.~Rota, and A.~Ratier,
  ``An introduction to meteosat second generation (msg),'' \emph{Bulletin of
  the American Meteorological Society}, vol.~83, no.~7, pp. 977--992, 2002.

\end{thebibliography}
}

\end{document}